%% file: main.tex
\documentclass{article}

\usepackage{xcolor}
\usepackage{color}
\usepackage{arxiv}
\usepackage{CJKutf8}
\usepackage[utf8]{inputenc} 
\usepackage[T1]{fontenc}    
\usepackage{hyperref}       
\usepackage{url}            
\usepackage{booktabs}       
\usepackage{amsfonts}       
\usepackage{nicefrac}       
\usepackage{microtype}      
\usepackage{lipsum}
\usepackage{multirow}
\usepackage{subfigure}
\usepackage{graphicx}
\usepackage{footnote}
\usepackage{amsmath,amsfonts,amssymb}
\usepackage{threeparttable}
\usepackage{CJK}
\usepackage{longtable}

\usepackage[linesnumbered,ruled,vlined,commentsnumbered]{algorithm2e}
\makesavenoteenv{tabular}
\makesavenoteenv{table}

\usepackage{subfiles} 
\usepackage{makecell}

\usepackage{array}

\newcommand{\MODEL}{PanGu-$\Sigma$}

\usepackage[normalem]{ulem}

\title{\MODEL: Towards Trillion Parameter Language Model with Sparse Heterogeneous Computing}

\author{Xiaozhe Ren$^{1}\thanks{Equal Contribution}$\\ \and \textbf{Pingyi Zhou$^{1*}$}\\ \and \textbf{Xinfan Meng$^{1*}$}\\ \and \textbf{Xinjing Huang$^{2*}$} \and \textbf{Yadao Wang$^{1*}$} \and \textbf{Weichao Wang$^1$} \and \textbf{Pengfei Li$^1$} \and \textbf{Xiaoda Zhang$^2$} \\ \and \textbf{Alexander Podolskiy$^1$} \and \textbf{Grigory Arshinov $^1$} \and \textbf{Andrey Bout $^1$} \and \textbf{Irina Piontkovskaya $^1$} \and \textbf{Jiansheng Wei$^1$} \and \textbf{Xin Jiang$^1$} \\ \and \textbf{Teng Su$^2$~~~~~~~~Qun Liu$^1$~~~~~~~~Jun Yao$^1$} \\ \\
\textbf{$^1$Noah's Ark Lab, Huawei Technologies} \\
\textbf{$^2$Distributed and Parallel Software Lab, Huawei Technologies}
}

\begin{document}
\date{}
\maketitle

\begin{abstract}
The scaling of large language models has greatly improved natural language understanding, generation, and reasoning. In this work, we develop a system that trained a trillion-parameter language model on a cluster of Ascend 910 AI processors~\footnote{\url{https://e.huawei.com/en/products/servers/ascend}} and MindSpore framework~\footnote{\url{https://www.mindspore.cn/en}}, and present the language model with 1.085T parameters named \MODEL. With parameter inherent from PanGu-$\alpha$~\cite{Zeng2021PanGuLA}, we extend the dense Transformer model to sparse one with \textit{Random Routed Experts} (RRE), and efficiently train the model over 329B tokens by 
using \textit{Expert Computation and Storage Separation} (ECSS). This resulted in a 6.3x increase in training throughput through heterogeneous computing. Our experimental findings show that \MODEL\ provides state-of-the-art performance in zero-shot learning of various Chinese NLP downstream tasks. Moreover, it demonstrates strong abilities when fine-tuned in application data of open-domain dialogue, question answering, machine translation and code generation.

\end{abstract}


\section{Introduction}\label{sec:introduction}

\subfile{sections/introduction}

\section{Model}\label{sec:model}

\subfile{sections/model}

\section{Dataset}\label{sec:dataset}

\subfile{sections/dataset}

\section{System}\label{sec:system}

\subfile{sections/system}

\section{Experiments}\label{sec:experiments}

\subfile{sections/experiments2}

\section{Conclusion and Future Work}\label{sec:conclusion}

\subfile{sections/future}

\section{Acknowledgements}
We would like to thank Bin Zhou, Zhiwei Wang, Yasheng Wang, Liangyou Li, Bin He and Fanyi Du for their great support for this work; Chen Li, Yifan Yao, Kaisheng Wang, Zhenzhang Yang, Zhongzhe Hu,  Zhepeng Sun, Zhijian Guo, Jun Wang, and Ziqiang Chen for their help to handle infrastructure issues.  

\clearpage

\bibliographystyle{unsrt}  
\bibliography{references}  

\appendix

\subfile{appendices/multiple_choice}

\end{document}

%% file: sections/Introduction.tex
Large Language Models (LLMs)~\cite[etc.]{brown2020GPT3,2020Exploring, Zeng2021PanGuLA,2022PaLM,2021Scaling,Zhang2022OPTOP,CPM2020,ERNIE3TITAN,zeng2022glm,scao2022bloom} have demonstrated unprecedented capabilities and potential in the areas of natural language understanding, generation and reasoning. By utilizing vast amount of textual data, the performance of language models scales up with compute budget and model parameters, demonstrating strong zero/few-shot learning abilities or even emergence abilities~\cite{2022PaLM,wei2022emergent}. Several large language models with hundreds of billion parameters have been published since GPT-3~\cite{brown2020GPT3}, including but not limited to Megatron-Turing NLG~\cite{smith2022using}, PanGu-$\alpha$~\cite{Zeng2021PanGuLA}, ERNIE 3.0 Titan~\cite{ERNIE3TITAN}, Gopher~\cite{2021Scaling}, PaLM~\cite{2022PaLM}, OPT~\cite{Zhang2022OPTOP}, 
Bloom~\cite{scao2022bloom}, and GLM-130B~\cite{zeng2022glm}. Researchers start to build even larger language models with more than one trillion parameters. Typically, this is accomplished by leveraging sparsely-activated models such as Mixture-of-Experts (MoE)~\cite{shazeer2017outrageously}. Among the trillion-parameter models currently in existence, there are several noteworthy work such as Switch-C~\cite{fedus2021switch}, GLaM~\cite{2021GLaM},  MoE-1.1T~\cite{artetxe2021efficient}, Wu Dao 2.0~\cite{wudao2}, and M6-10T~\cite{lin2021m6}. However, only a select few have published comprehensive evaluation results over a wide range of tasks while simultaneously achieving anticipated performance. In our experience, the primary difficulty lies in the scaling efficiency.

Recent studies on the scaling laws of language models~\cite{2020Scaling,2022Training,clark2022unified} demonstrate the necessity of training LLMs with sufficient amount of training data and corresponding compute budget to achieve optimal performance. Therefore, one of the main motivation for this work is to design a scalable model architecture and an efficient distributed training system that can consume the data with high training throughput.

\begin{itemize}
\item \textbf{Model Scaling}.
Model performance of LLMs is expected to scale up with larger model size. Comparing to the expensive computational cost for training dense Transformer model, sparse architectures such as Mixture-of-Experts (MoE)~\cite{shazeer2017outrageously,fedus2021switch,2021GLaM,2021GShard} are considered to be an appealing choice to scale model size up without incuring linear increase in computational cost. However, MoE models suffer from the problems such as unbalanced workload and all-to-all communication latency. Moreover, how to extend existing dense model with MoE and how many experts to allocate in each layer remain open problems. Therefore, designing a trillion parameter sparse model with high performance and training efficiency is a significant yet challenging task.

\item \textbf{System Scaling}.
Frameworks such as DeepSpeed~\footnote{\url{https://www.deepspeed.ai/}} have been proposed to support training trillion parameter models. In practice, the main barrier often lies on limited compute budget, or more specifically the number of accelerating devices (e.g., GPU, NPU, TPU) that can be used. By utilizing techniques such as tensor parallelism~\cite{Megatron}, pipeline parallelism~\cite{Gpipe}, zero redundancy optimizer~\cite{zero} and rematerialization~\cite{2016Training}, practitioners can train trillion-parameter model with feasible batch sizes across thousands of accelerating devices. Alternatively, practitioners can reduce the amount of computation resources by utilizing heterogeneous computing techniques such as offloading some of the computation to host devices~\cite{2021ZeRO}. However, the current techniques inevitably hinder the training throughput due to slow bandwidth between the host and device as well as weak computing capabilities of CPUs compared to accelerating devices, which prevent feeding large language models with reasonably amount of data and achieving optimal performance. Therefore, how to efficiently scale the system performance with limited computation budget is critical to the performance of large language models.
\end{itemize}

In this work, we present \MODEL\ , a large language model with sparse architecture containing 1.085 trillion parameters. We develop \MODEL\ model under the framework of MindSpore~\footnote{\url{https://gitee.com/mindspore/mindspore}} and train it on a cluster with only 512 Ascend 910 AI Accelerators~\cite{2021Ascend} with 329 billion tokens over 100 days. \MODEL\ inherent parameters from PanGu-$\alpha$~\cite{Zeng2021PanGuLA} with Transformer decoder architecture and are extended via Random Routed Experts (RRE). Different from conventional MoE, RRE adopts two-level routing. At the first level experts are grouped by domain or task, and at the second level tokens are randomly and uniformly mapped to experts in each group without using any learnable gating function as in MoE. With the design of RRE, one can easily extract sub-models from the \MODEL\ for various downstream applications such as dialogue, translation, code generation or general nature language understanding. To make training system efficient and scalable, we propose Expert Computation and Storage Separation (ECSS) mechanism, which achieves 69905 tokens/s observed throughput in training 1.085 trillion \MODEL\ on cluster of 512 Ascend 910 accelerators, and reduces Host-to-Device and Device-to-Host communication as well as optimizer update computation by a large margin. As a whole, the training throughput is improved by 6.3x compared to the model of the same hyper-parameters but with MoE architecture. By consuming 329B tokens in more than 40 natural and programming languages, the sub-modal of \MODEL\ in Chinese domain significantly outperforms the previous SOTA models including PanGu-$\alpha$ with 13B parameters and ERNIE 3.0 Titan~\cite{ERNIE3TITAN} with 260B parameters over 16 downstream tasks in six categories in the zero-shot setting without any multitask finetuning or instruction tuning. We also test the performance of fine-tuned \MODEL\ on several applications domain such as dialogue, machine translation and code generation. \MODEL\ outperforms the SOTA models in the corresponding areas.

The rest of the technical report is organized as follows. Section 2 introduces the design philosophy and the architecture of \MODEL\ model. Section 3 introduces the collection and organization of the dataset. Section 4 describes system design and acceleration techniques. Section 5 presents the experimental results of \MODEL\ model.

%% file: sections/Model.tex

\subsection{Design Principles}

\MODEL\ aims to achieve the following goals.
\begin{itemize}
  \item \textbf{Performance}: state-of-the-art NLP performance across multiple domains and tasks.
  \item \textbf{Efficiency}: training trillion parameters model with maximum system performance on a modest cluster.
  \item \textbf{Usability}: extendable to various domains or tasks, without need of retraining the model from scratch.
  \item \textbf{Deployment}: easily customizable and deployable in various real-world settings.
\end{itemize}

Achieving all the above goals at the same time is very challenging. Considering the first goal, a language model that can generalize and perform well across domains should have a very large number of parameters and be trained on large amount of data according to the scaling law~\cite{2020Scaling,2022Training,clark2022unified}. However, training such a large model also means that a high-end cluster is mandatory, which somehow contradicts with the second goal. And the larger scale of the model also leads to increasing cost in deploying the trained model, which is related to the fourth goal.

Considering the high computational cost incurring during the training phase, we want the resulted model to be practically usable and efficient in many real applications. With this goal in mind, we propose to train the model in multiple domains and make it further extendable to any number of domains in a continuous learning paradigm, subject to the computation resource.

During training phase, the trillion parameters \MODEL\ model is fed with data from multiple domains. However, in the deployment phase, it is often unnecessary or even impossible to host the trillion parameters model for every application. Therefore, a model that allows for the grouping and separation of its parameters based on various training and deployment setups offers significant advantages.

\subsection{PanGu-$\Sigma$ Architecture}

\subsubsection{Overview}

\begin{figure*}[!ht]
    \centering
    \includegraphics[width=0.75\textwidth]{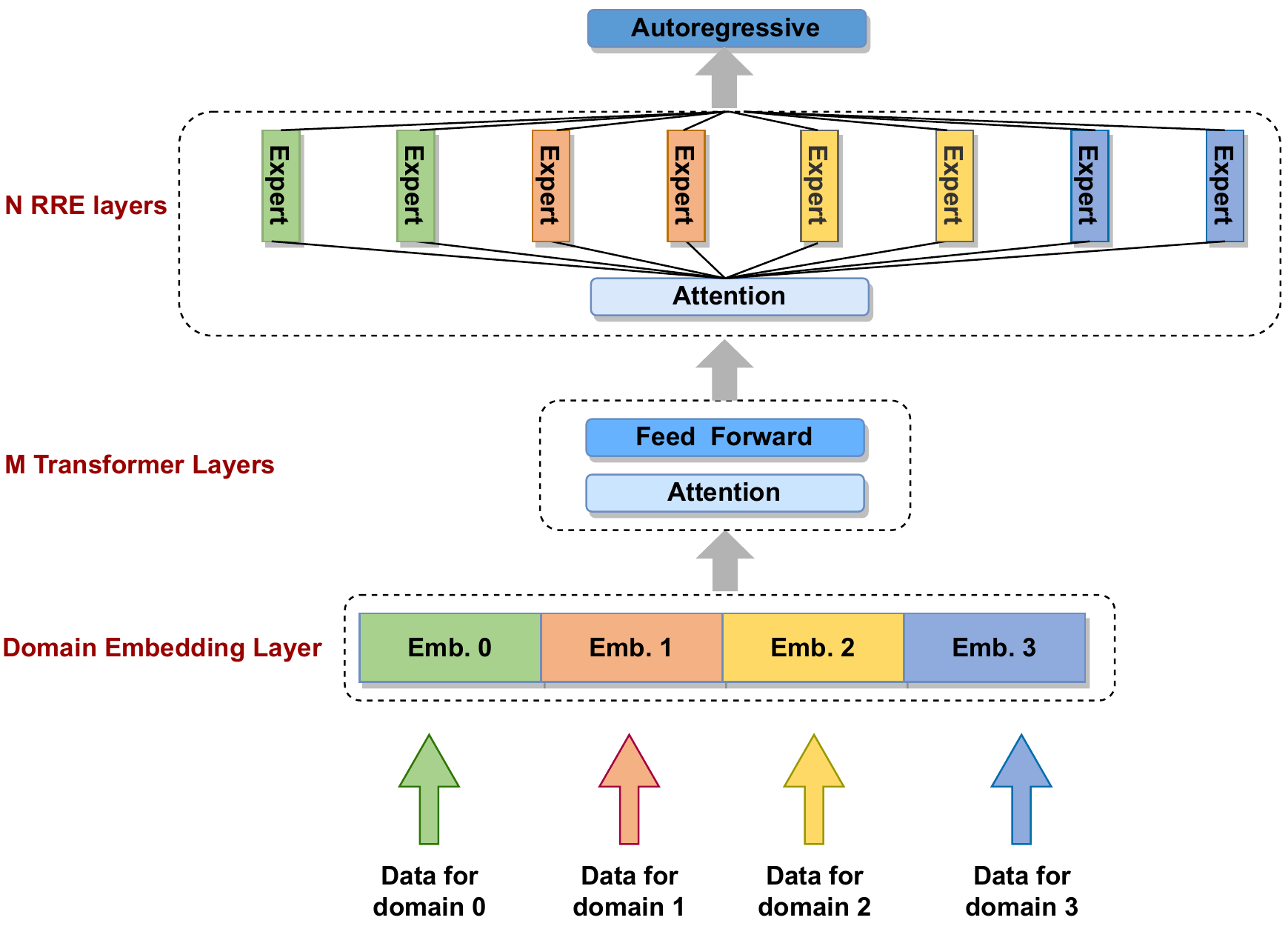}
    \caption{PanGu-$\Sigma$ architecture. The architecture is mixed by dense transformer layers and sparse transformer layers. The lower M layers are dense layers shared across different domains. The upper N transformer layers' feed-forward part are sparsely activated via Random Routed Experts (RRE). Tokens from different domains have different embeddings. }
    \label{fig:alpha_to_sigma}
\end{figure*}

\MODEL\ adopts an auto-regressive language modeling with stacked transformer decoder layers and a query layer on the top. The \MODEL\  architecture offers a flexible design. The bottom M layers are globally shared across all the domains, and the top N layers (including the query layer) are sparsely activated according to the domains of the input data. In each RRE layers, there are K experts in G groups in total, the number of experts in each group can be different. 
This flexible design offers three mode.

\begin{itemize}
  \item \textbf{Mixed mode}: when $M > 0$, $N > 0$ and $K > 0$, model contains both sparse RRE layers and dense layers. 
  \item \textbf{Dense mode}: when $N = 0$ or $K = 1$, the architecture will reduce to a dense PanGu-$\alpha$ model.
  \item \textbf{Sparse mode}: when $M=0$ and $K > 1$, the architecture will be a sparse model.
\end{itemize}
 
 In this trillion-parameters modeling practice, We use the mixed configuration by placing the shared parameters close to the input layer (bottom) and all the sparsely activated expert parameters close to the output layer (top). In the model designing stage, we benchmark various experts placement strategies on smaller scale models and the selected strategy obtains the lowest language modeling perplexity. Our hypothesis is that bottom layers tends to learn general knowledge, while the specific knowledge is in a higher level of abstraction and is more appropriate to be learned by the top layers. In the token embedding layer, we choose to use different embedding matrices for different domains.

\subsubsection{Random Routed Experts}

In the top N layers, we replace each feed-forward sub-layer with multiple conditionally activated feed-forward sub-layers (experts), following the Mixture of Experts (MoE) paradigm.

A key question in designing MoE architecture is how to route tokens to experts. For \MODEL\ , we propose a Random Routed Experts (RRE) mechanism, which is inspired by Hash Layers proposed in~\cite{2021Hash}.
Specifically, RRE routes the tokens by IDs in a two-level manner. In the first level, the token is mapped to a group of candidate experts by domain, and then in the second level, one expert in this group is chosen according to a token-expert routing map to process the token. The routing map is randomly initialized and each layer has a independently initialized mapping for balancing the computation.

\begin{figure*}[!ht]
    \centering
    \includegraphics[width=0.7\textwidth]{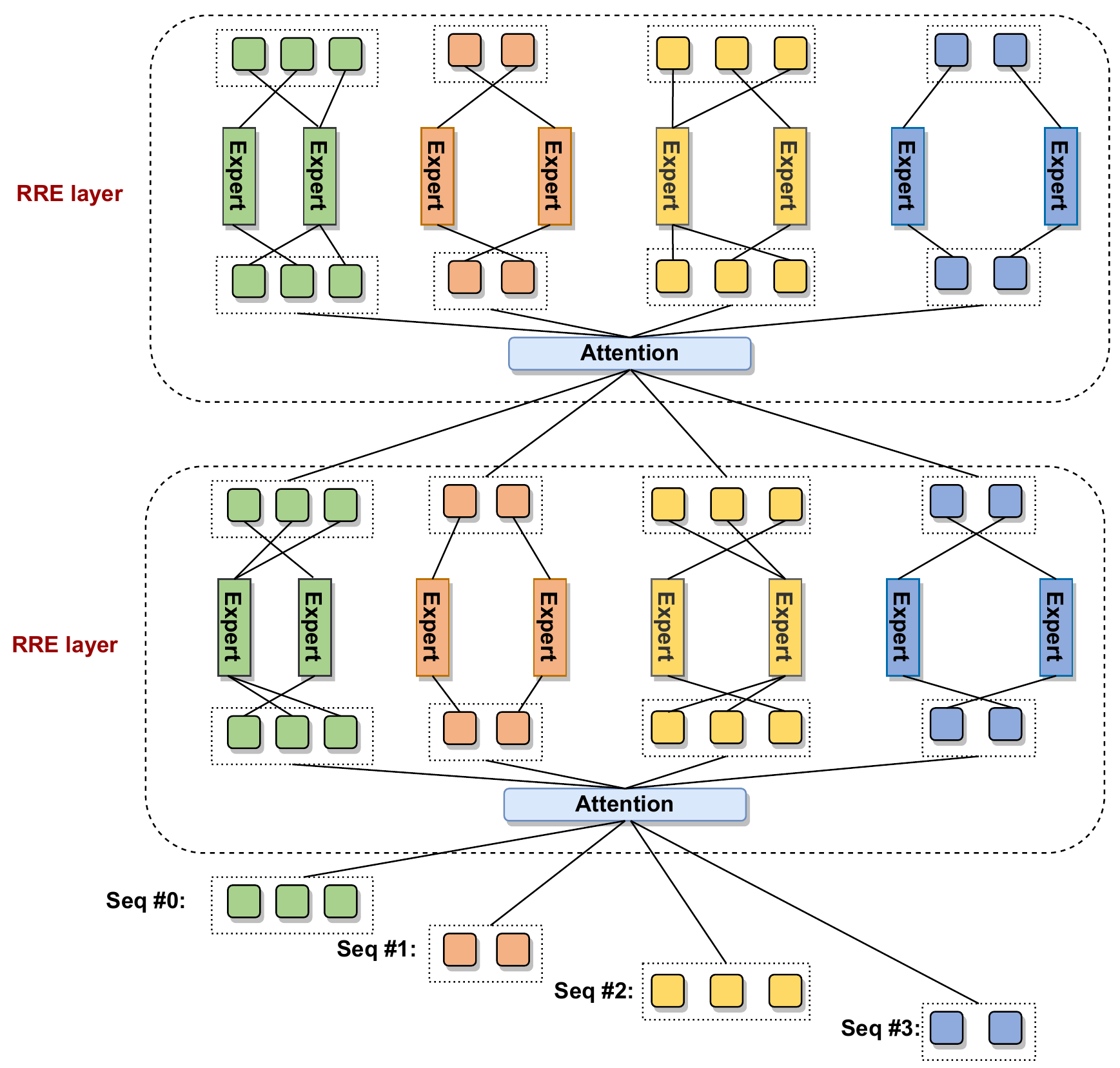}
    \caption{Random Routed Experts (RRE) in \MODEL\ . The token is first routed to a group of experts by domain, and then randomly routed to one of the experts in that domain. There is no learnable routers in the model.}
    \label{fig:alpha_to_sigma}
\end{figure*}

RRE has several advantages over the commonly-used learnable routers.

\begin{itemize}
    \item During training, \MODEL\ allows for the addition, modification, or removal of domain-specific experts without any impact on the other experts. This attribute makes \MODEL\ highly flexible for alleviating the commonly encountered problem of catastrophic forgetting, which is crucial for life-long or continual learning.
    \item In most real-world deployment setting, it is unnecessary or impractical to deploy a trillion-parameter model. \MODEL\ allows one to extract a sub-model for specific domains according to practical requirements and only deploy the sub-model. The sub-model may contain tens of billion parameters but still keep the predictive power of the original model on the target domains. Using this extract-and-deploy operation, we can easily deploy models for multiple industrial applications.
    \item All the conventional MoE models rely on all-to-all communication collective operation to move data between experts residing on different devices. With our proposed two-level routing, experts from different domains don't exchange tokens, and all-to-all communication is constrained \emph{within} each domain. As a result, the expensive global all-to-all operation is reduced to grouped all-to-all, saving much communication volume and reducing the end-to-end training latency.
    \item Learnable router needs more computation, and can suffer from problem of unbalanced loads across the experts, which typically makes the training process less stable. RRE avoids all the above pitfalls since no additional parameters are introduced and randomly initialized routing table helps to balance the loads on experts.
\end{itemize}

RRE requires a routing map which is initialized before pretraining, Algorithm 1 describes how we construct the routing table. 

\begin{algorithm}[H]
\caption{Routing table construction procedure in Random Routed Experts (RRE) mechanism.}
 \SetKwInOut{Input}{input}\SetKwInOut{Output}{output}
 \Input{number of domain $d$, number of layers $l$, number of experts per domain per layer $e$, size of vocabulary $V$.}
 \Output{$T$, a tensor of shape $(d, l, V)$ acting as the RRE routing table.}
 set random seed to 0 \;
 initialize $T$ \;
 initialize $\vec{u}$, a vector $[0, ..., V-1]$ of size $V$ \;
 $\tilde{V} = \lfloor V / e \rfloor \cdot e$ \;
 $\vec{u}[0: \tilde{V}-1] = \lfloor\vec{u}[0: \tilde{V}-1] / e  \rfloor$ \;
 $\vec{u}[\tilde{V}: V-1] = [0, ..., V -\tilde{V} -1 ] $ \;
  
 \For{$j \leftarrow 0 $ \KwTo $l - 1$}{
 \For{$i \leftarrow 0 $ \KwTo $d - 1 $}{
    shuffle $\vec{v}$, a vector $[0, ..., V-1]$ of size $V$ \;
    $T[i][j][\vec{v}] = \vec{u} + i * e $ \;
  }
 }
\end{algorithm}

%% file: sections/Dataset.tex
\subsection{Collection}
To better demonstrate the capability of \MODEL\ model to efficiently and independently learn from multiple domains, we collect datasets in 40 domains, with a large amount of data in four major domains: Chinese, English, Bilingual (Chinese and English) and code. The remaining domains with smaller portion consists of 26 other monolingual natural languages, 6 programming languages, and textual data from finance, health, law, and poetry domains, respectively. 

For Chinese texts, we collect the WuDaoCorpora 2.0~\cite{Yuan2021WuDaoCorporaAS} which contains 200GB and the CLUECorpus2020~\cite{Xu2020CLUECorpus2020AL} which contains 100GB. For English texts, the Pile dataset~\cite{Gao2021ThePA} which contains 800GB and C4 dataset~\cite{2020Exploring} which contains 750GB were collected. For code, we use the Python code (147GB) which has been used in PanGu-Coder~\cite{Christopoulou2022PanGuCoderPS}, as well as the Java code (161GB) from GHTorrent~\cite{Gousios2013TheGD} , which are then filtered by file size ($<$1MB), average number of characters per line ($<$200), maximum number of characters per line ($<$1000) and their compilablity. Then, these collected English, Chinese and code texts data was sampled and distributed to the four major domains. Finally, we get more than 300B tokens for the four major domains. The detailed statistics of data distribution and data sources in four major domains are presented in Table~\ref{tab:data_dis}.

For the remaining 36 domains, the data for 26 monolingual domains are mainly from CCAligned~\cite{ElKishky2019AMC} and CCMatrix~\cite{Schwenk2019CCMatrixMB}. Similar to the code domain mentioned above, the data for 6 programming language domains are collected through GHTorrent~\cite{Gousios2013TheGD} and filtered in the similar way. Finance domain data is filtered from the WuDaoCorpora 2.0~\cite{Yuan2021WuDaoCorporaAS} using the tags. Health domain data is from Chinese MedDialog Dataset ~\cite{Zeng2020MedDialogLM}. Law domain data is sampled from CAIL2018~\cite{Xiao2018CAIL2018AL}. Poetry domain dataset is from Werneror-Poetery~\footnote{\url{https://github.com/Werneror/Poetry}}. Finally, we sampled more than 25B tokens for the 36 domains.

\begin{table}
\centering
\caption{Data distribution and data sources in four main domains}
\label{tab:data_dis}
\scalebox{0.8}{
\begin{tabular}{cccc} 
\hline
Domain ID & Domain                                                                & Tokens (Billion)                                                             & Data source                                                                                                    \\ 
\hline
0         & \begin{tabular}[c]{@{}c@{}}\textbf{Bilingual}\\(Chinese, English)\end{tabular}      & \begin{tabular}[c]{@{}c@{}}77.51 B\\Chinese (38.75) + English(38.76B)\end{tabular} & CLUECorpus2020 , C4                                                                                            \\ 
\hline
1         & \textbf{\textbf{Chinese}}                                             & 75.47 B                                                                            & WuDaoCorpora 2.0~ ~                                                                                            \\ 
\hline
2         & \textbf{\textbf{English}}                                             & 75.90 B                                                                            & Pile , C4                                                                                                      \\ 
\hline
3         & \begin{tabular}[c]{@{}c@{}}\textbf{Code}\\(Python, Java)\end{tabular} & \begin{tabular}[c]{@{}c@{}}75.24 B\\Python (50.24B) + Java (25B)~~~~~\end{tabular} & \begin{tabular}[c]{@{}c@{}}Python (PanGu-Coder)~\\Java (GHTorrent)\end{tabular}  \\ 
\hline
\end{tabular}
}
\end{table}

\subsection{Format}
For the four major domains, each can be adapted to different downstream tasks. In order to better support domain-specific downstream tasks, this paper uses different data format for different domains. For Chinese and English domains, the <EOT> token which indicates the end of training text is inserted at the end of each training sample.

\begin{figure*}[!ht]
    \centering
    \includegraphics[width=0.8\textwidth]{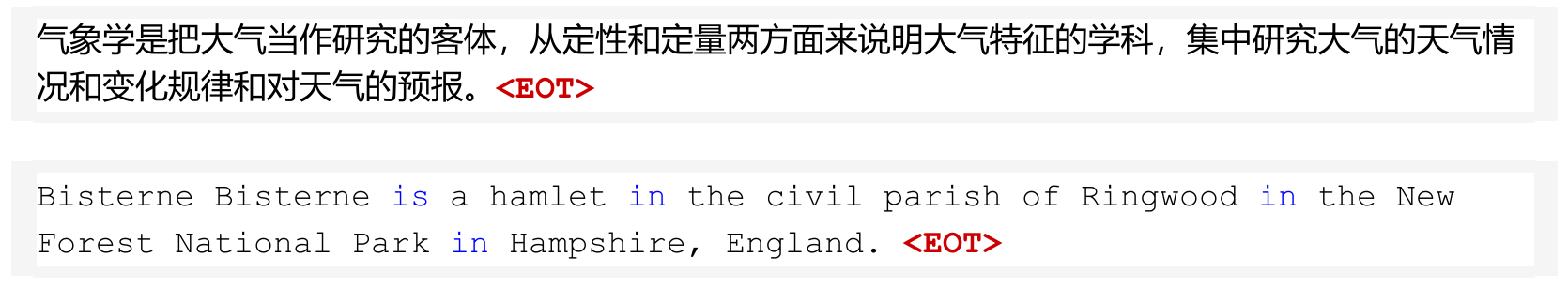}
    \caption{Data format of Chinese and English domains.
    }
    \label{fig:sample_1}
\end{figure*}

For Bilingual domain, the <EN> or <CN> token is inserted into the head of the training sample according to the source of the training sample (either from the Chinese dataset or the English dataset), and the <EOT> token is inserted at the end of each training sample.

\begin{figure*}[!ht]
    \centering
    \includegraphics[width=0.8\textwidth]{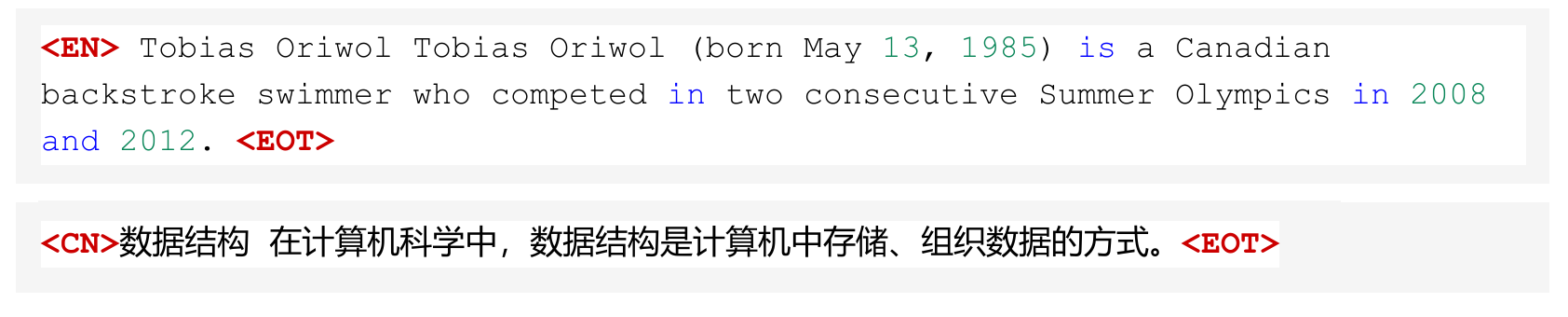}
    \caption{Data format of Bilingual domain.
    }
    \label{fig:sample_2}
\end{figure*}

For the code domain, the <Python> or <Java> token is inserted into the head of the training sample based on the programming language type of the training sample, and the <EOT> token is inserted at the end of each training sample.

For the remaining 36 domains, the data formats of 26 monolingual domains, finance, health, law, and poetry domains are the same as the Chinese and English domains, and the data format of 6 programming language domains is the same as the code domain.

\begin{figure*}[!ht]
    \centering
    \includegraphics[width=0.8\textwidth]{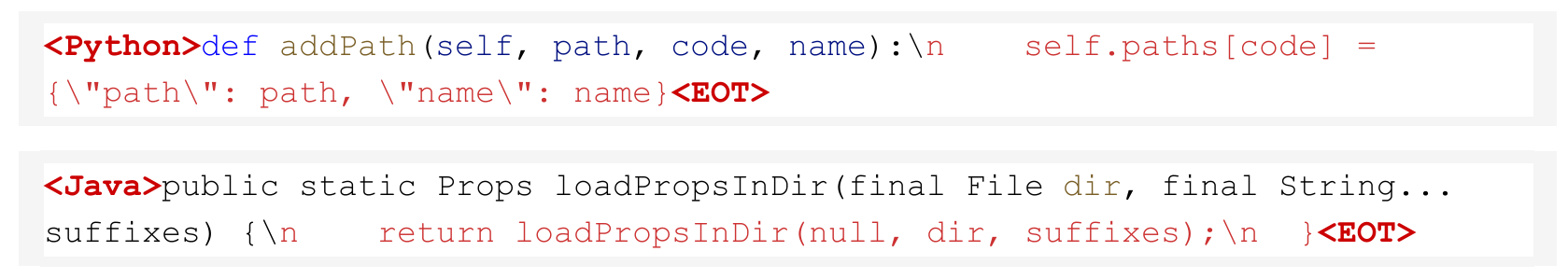}
    \caption{Data format of Code domain.
    }
    \label{fig:sample_3}
\end{figure*}

\begin{figure*}[!ht]
    \centering
    \includegraphics[width=0.7\textwidth]{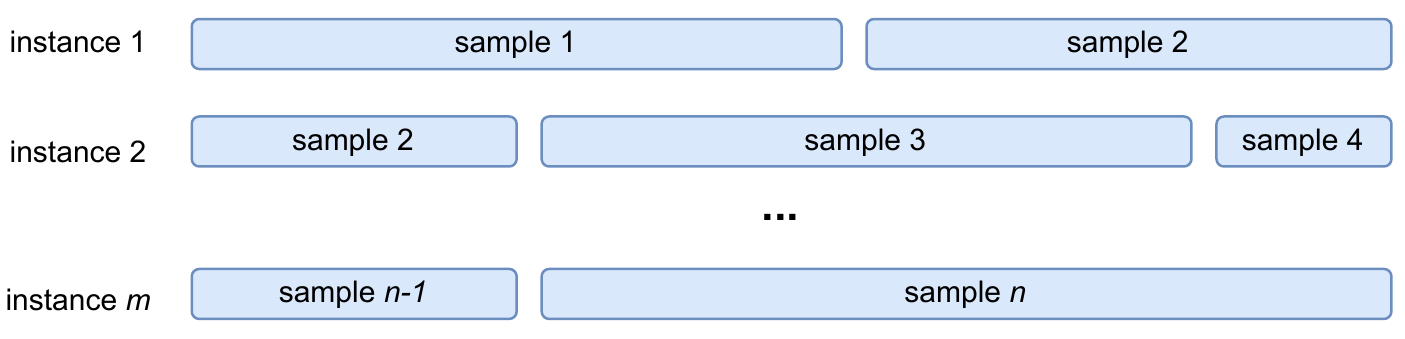}
    \caption{Input format during model pre-training.
    }
    \label{fig:format_1}
\end{figure*}

\begin{figure*}[!ht]
    \centering
    \includegraphics[width=0.7\textwidth]{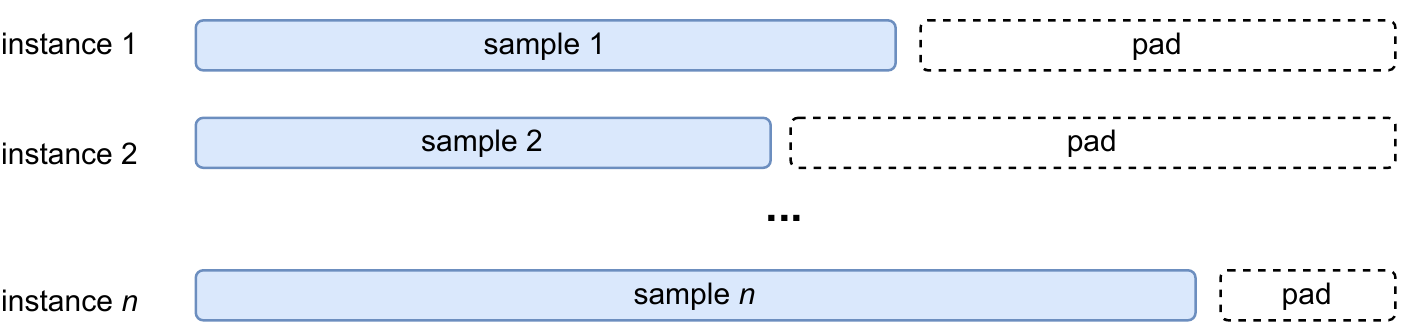}
    \caption{Input format during model fine-tuning.
    }
    \label{fig:format_2}
\end{figure*}

For a formatted data set $D$, suppose it contains n training samples $D=\left \{ s_{1}, s_{2}, \dots,  s_{n} \right \} $. To make full use of the computing power of the Ascend 910 cluster and accelerate training in the pre-training phase, we concatenate all samples in the data set into a sequence, and then intercept training instances in the concatenated sequence according to the fixed length (1024), as shown in Figure~\ref{fig:format_1}. In the fine-tune phase, for each training sample in the formatted dataset, if the length is less than the fixed length, we pad the sample to the fixed length with a special token <Pad>. If the length is greater than the fixed length, the extra part is truncated. Figure~\ref{fig:format_2} shows the process. Different to 
PanGu-$\alpha$ model, each training sample of \MODEL\ model contains two field: input sequence of token IDs which are training instance and their domain ID. The domain ID indicates which domain the training instance belongs to. The RRE layers of the \MODEL\ model decide which experts the training tokens is routed to by the domain ID. 

%% file: sections/System.tex
\MODEL\ is implemented with MindSpore 1.6 framework~\footnote{\url{https://www.mindspore.cn/versions/en}} and trained on 512 Ascend 910 accelerators (also know as Ascend 910 NPU).

Training a trillion parameters language model poses multiple challenges. First, it requires enormous amount of memory in training. Although the sparse architecture can effectively save computation, it doesn't reduce the memory consumption and we still need to store all the parameters and optimization states inside the accelerator memory. Assuming Adam optimizer~\cite{2014Adam} with mixed-precision training~\cite{micikevicius2018mixed} is used, a 1T model typically consumes 16TB memory in total just for parameters, gradients and optimizer states. During training, the model needs extra memory for input data, network activations, communication buffers and temporary variables. We estimate that training a \MODEL\ model with 1 trillion parameters with a reasonably batch size needs more than 32TB memory and requires more than 1,000 Ascend 910 accelerators or NVIDIA V100 GPUs with 32GB High Bandwidth Memory (HBM). 

Instead of pouring lots of hardware resources to scale-up the model, we aim to train \MODEL\ with a reasonably-sized cluster of 512 Ascend accelerators. To this end, we adopt the heterogeneous training and offload the optimizer states to CPU\cite{2021ZeRO}. After enabling heterogeneous training, all optimizer states are moved from accelerator to the host with 750GB host memory and KunPeng 920 CPU~\footnote{\url{https://www.hisilicon.com/en/products/Kunpeng/Huawei-Kunpeng/Huawei-Kunpeng-920}}, and we can fit the entire training process into the cluster.

\begin{figure*}[htb]
    \centering
    \includegraphics[width=1.05\textwidth]{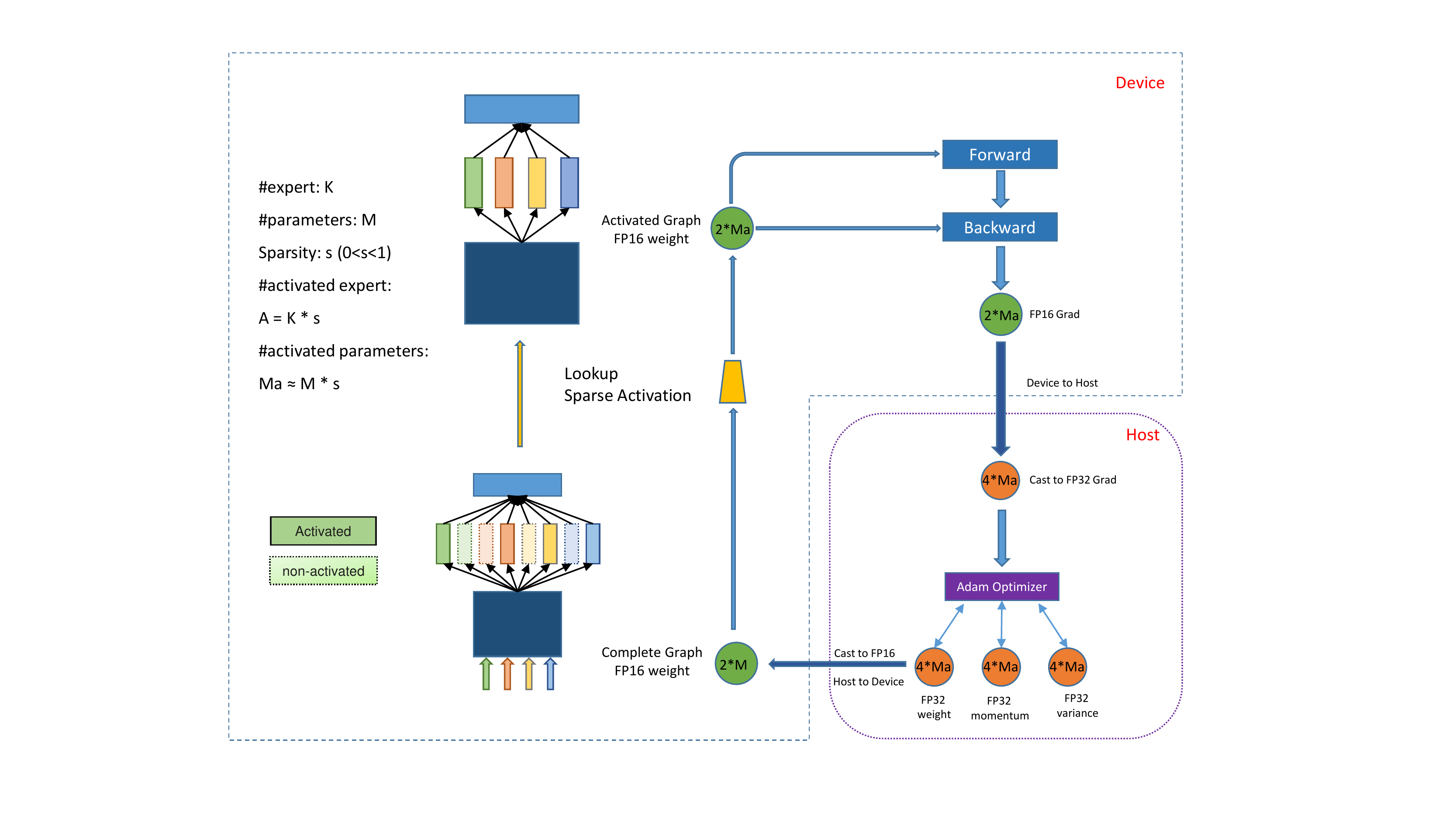}
    \caption{Expert Computation and Storage Separation (ECSS) in traning \MODEL\ . In each iteration, with sparsity ratio $s\in(0,1]$ and expert amount K, only number of $A=Ks$ experts activated by lookup operation, which reduce the communication cost between device and host, and cost of forward and backward computation in device as well as optimizer operations in host.}
    \label{fig:ecss}
\end{figure*}

Second, the system throughput is unacceptable after enabling vanilla optimizer offloading. The root cause is again the sheer amount of parameters.  Gradients and updated parameters need to be exchanged via the slow host-to-device and device-to-host communication, and CPUs need to iterate thorough all parameters and update them. To improve the training throughput, we leverage the sparse nature of \MODEL\ architecture. Since \MODEL\ use a sparse architecture and most of its parameters are conditionally activated, the optimizer only need to update part of experts in one iteration. So we propose \textbf{Expert Computation and Storage Separation} (ECSS) method as illustrated in Figure~\ref{fig:ecss}.

In Expert Computation and Storage Separation, we consider experts as knowledge database to store specific knowledge of different tasks or domains. In each iteration, experts are sparsely activated by different token IDs with specific domain. In MindSpore, we use lookup operator to select parts of activated experts, and sparsely update their parameters in the backward computation. In optimizer CPU offload computing, MindSpore copy FP16 parameters from host CPU to NPU, compute the gradients on NPU, move FP16 gradients from NPU to CPU, and compute optimizer states and update parameters in the host CPU. With a lower experts sparsity ratio such as $0.1$, the computation cost is only near 10$\%$ of full model. 

\begin{figure*}[htb]
    \centering
    \includegraphics[width=0.7\textwidth]{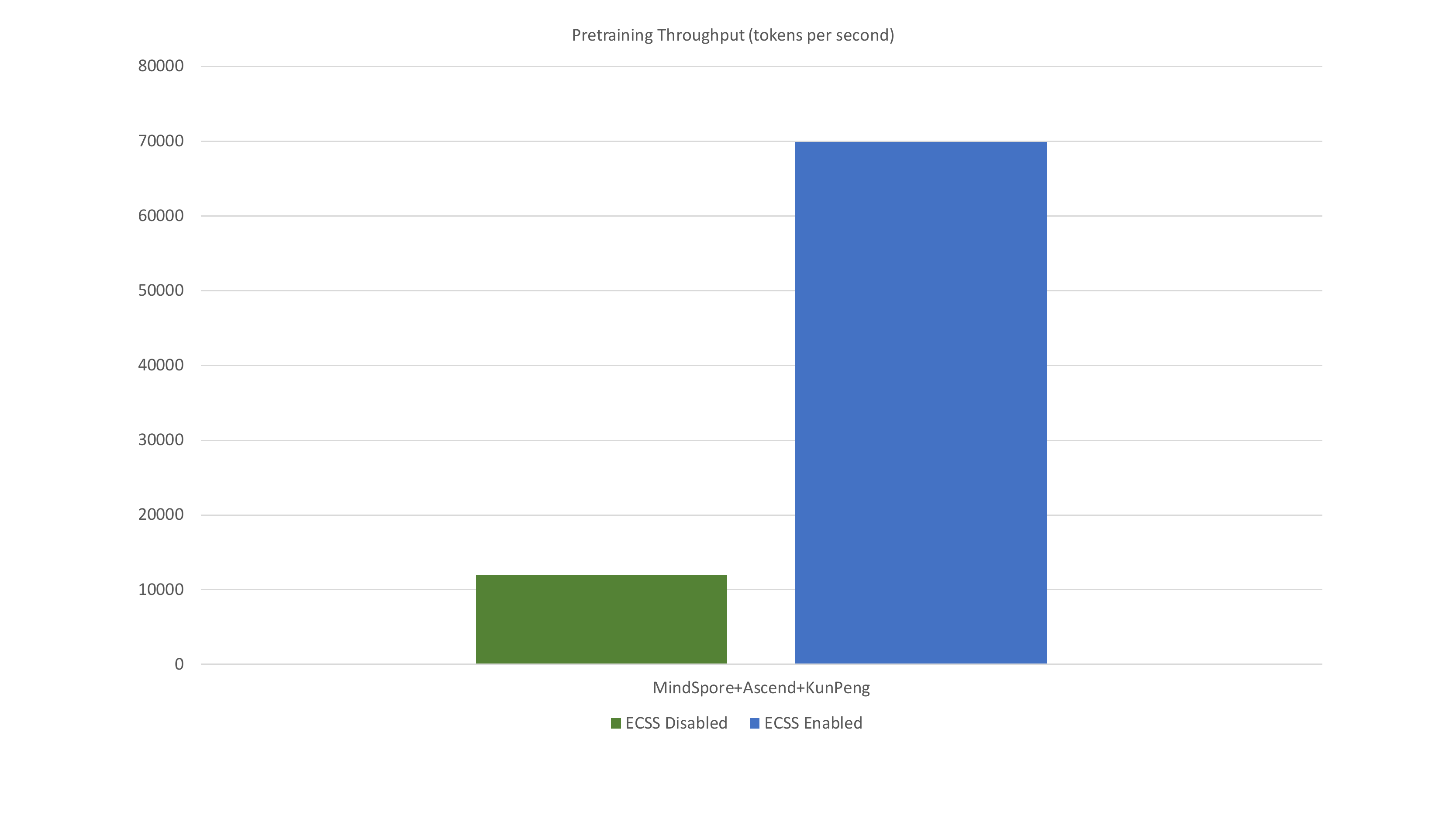}
    \caption{Training throughput (token/s) of \MODEL\ wo/w Expert Computation and Storage Separation (ECSS). ECSS can achieve 6.3x increase of training throughput.}
    \label{fig:throughput}
\end{figure*}

Besides ECSS with Ascend-KunPeng sparse heterogeneous computing, we also adopt other parallel training and accelerating techniques provided by MindSpore and CANN~\footnote{\url{https://www.hiascend.com/en/software/cann}}. We use 8-ways model parallel for all the attention and feed-forward layers, 64-ways expert parallel without replica and 64-ways data parallel for non-expert parts. To further optimize memory footprint, rematerialization~\cite{2016Training} and optimizer parallel~\cite{zero} are also adopted to reduce the peak memory consumption. We also use FastGelu and fused LayerNorm to accelerate point-wise computation. By combining all the techniques  together, we achieved 6.3 times throughput promotion compared to vanilla \MODEL\ heterogeneous training, as shown in Figure~\ref{fig:throughput}.



%% file: sections/Experiments2.tex
\subsection{Pretraining}

\subsubsection{Model Configuration}


We use the following \MODEL\ configuration for this work. The configuration mostly follows the 13B version of PanGu-$\alpha$ model. In this way, we can effective inherit the knowledge already learned by PanGu-$\alpha$. 

\begin{table}[ht]
\caption{Model Configuration}
\centering
\begin{tabular}{lllllll}
\hline
\#Layers (N+M) & \#Heads (N\_h) & Hidden size (d) & FFN size (d\_ff) & \#RRE layers (N) & \#Experts (K) & \# Groups (G)                        \\ \hline
 40       & 40 & 5120 & 20480 & 8 & 640 & 40                                            \\ \hline
\end{tabular}
\end{table}

\subsubsection{Pretraining settings}

We use a cluster of 64 nodes, with each node equipped with 8 Ascend 910 accelerators and MindSpore framework. High performance collective communication library Huawei Collective Communication Library (HCCL) is used to facilitate high speed high bandwidth communication for distributed training. 

\begin{figure*}[!ht]
    \centering
    \includegraphics[width=0.9\textwidth]{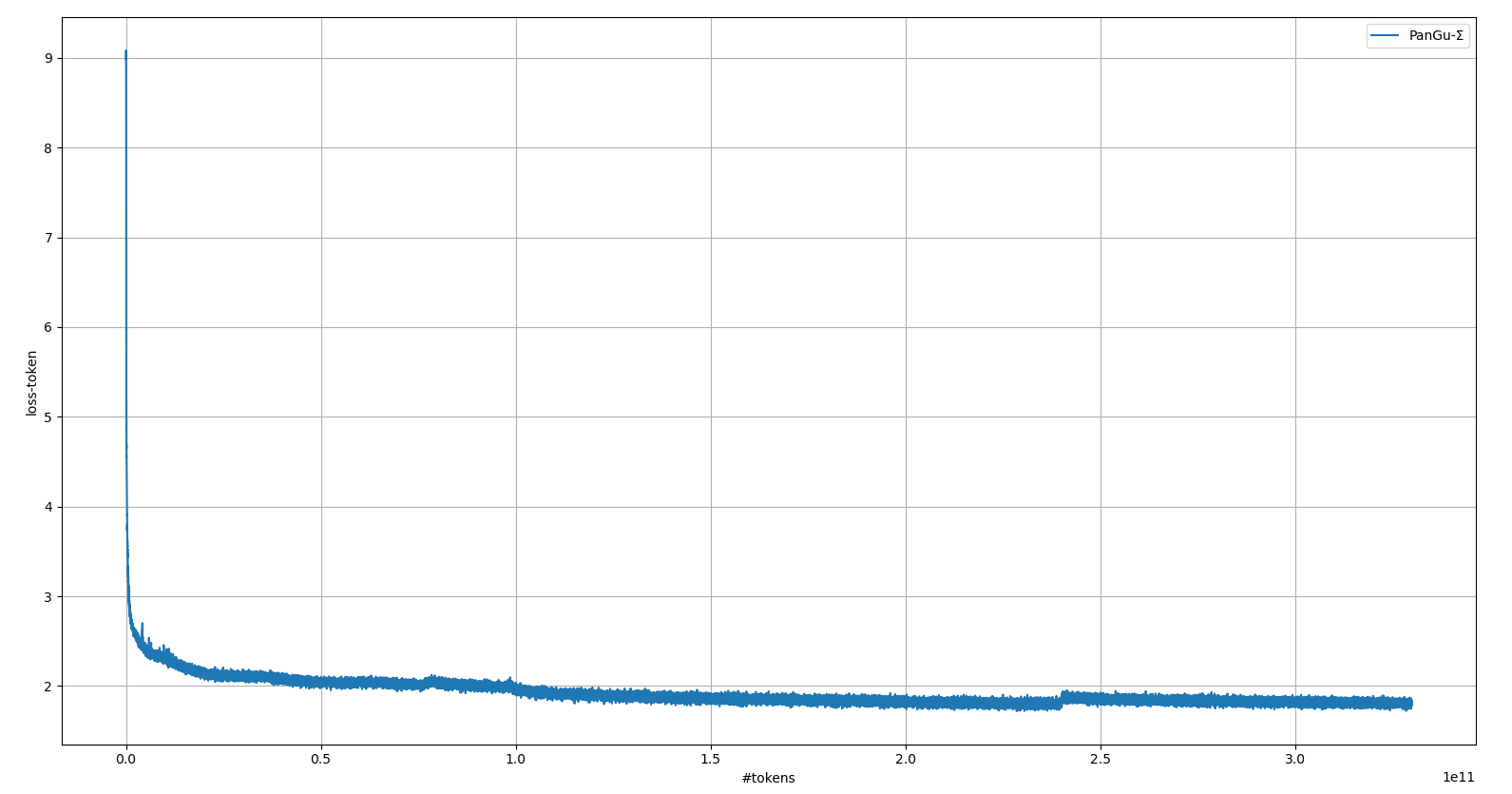}
    \caption{Pretraining Loss}
    \label{fig:loss}
\end{figure*}

\begin{figure*}[!ht]
    \centering
    \includegraphics[width=1\textwidth]{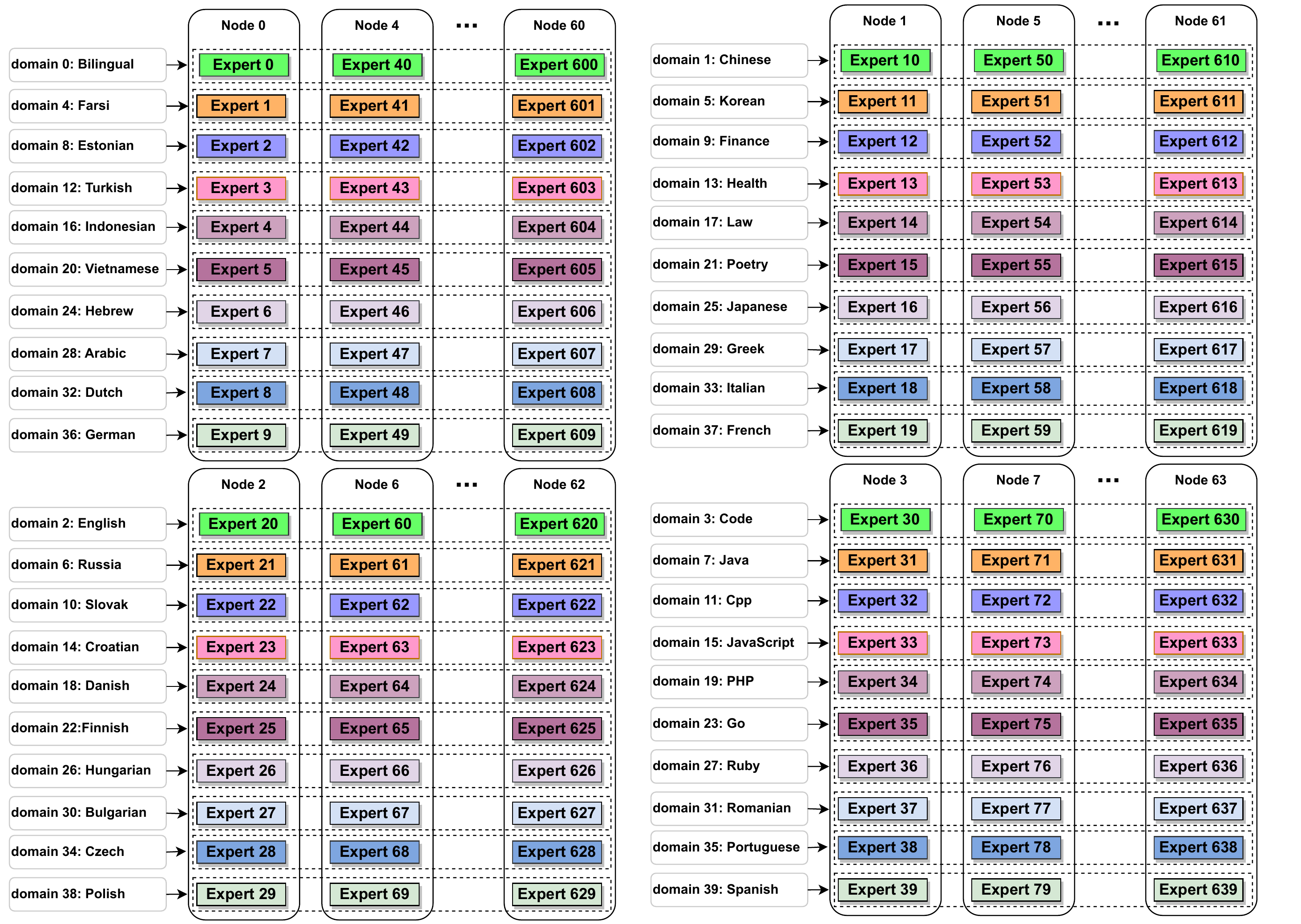}
    \caption{Mapping between domains and experts in \MODEL\ .The data from particular domains are routed to a group of experts lying across different devices. The color of experts distinguish their corresponding domains. There are ten experts from different domain on each device.}
    \label{fig:sigma_domain_to_expert}
\end{figure*}

There are two stages in \MODEL\ pretraining process. In the first stage, we activate four main domains' experts to consume data from all the four main domains including bilingual, Chinese, English and codes. In the second stage, we let all the experts to consume all domain's data. Figure~\ref{fig:sigma_domain_to_expert}  shows how 640 experts are assigned to 40 domain groups. We train \MODEL\ with global batch size of 512 with sequence length of 1024 for each sample. The pretraining lasts about 100 days. Figure~\ref{fig:loss} shows the loss curve of \MODEL\ pretraining.

Mixed-Precision training is enabled to speedup the training process. Apart from vocabulary embedding layer, loss function, Softmax operation, LayerNorm layers and Adam optimizer, all other operations adopt FP16 format.

Failure recovery is very important for long term large scale distributed training, especially for huge models like \MODEL\ . Therefore, a rigorous process of saving checkpoints and restarting from previous checkpoints is indispensable.  For a trillion-parameter model, one set of checkpoint storing all parameters and optimizer states for a single iteration already has a jaw-dropping 10TB size. Uploading checkpoints of such size to our long term object store is a challenging task, since uploading all checkpoints at the same time quickly saturates the network bandwidth and inevitably lead to training failure. To solve this issue, we launch the upload process in a round-robin style and limit the number of the simultaneously running process. This solution proves to be effective and stable for our entire training process.

\subsubsection{Hybrid Hyper-parameter ADAM Optimizer}

We design a Hybrid Hyper-parameter ADAM Optimizer to provides further stability for \MODEL\ during the pretraining phase.

To better understand \MODEL\  training process, we inspected the statistics of training states and find out that the gradients of RRE layers are much smaller than a non-sparse model. To tackle such a  problem, we first set a very small $\epsilon_1$ for all model parameters, then we go one step further and set an even smaller $\epsilon_2$ only for the RRE layers, since compared to the dense layers, sparse layers received smaller effective batch due to its conditionally-activated nature. Specifically, we set hybrid hyper-parameters for ADAM optimizer below:

\begin{table}[ht]
\caption{Hyper-parameters of \MODEL\ training.}
\centering
\begin{tabular}{lllllll}
\hline
$\beta_1$          & $\beta_2$ & $\epsilon_1$ & $\epsilon_2$ & end lr & warmup steps & decay steps                            \\ \hline

0.8          & 0.95       & 1e-8 & 1e-20 & 2e-5 & 5000 & 180000                                             \\ \hline
\end{tabular}
\end{table}


\subsection{Inheritance Learning}

To improve the training efficiency, accelerate model convergence, and reduce carbon emissions during training, the \MODEL\ model inherits the capabilities of the existing model, and then continues to train in four domains simultaneously. In this paper, \MODEL\ inherits the PanGu-$\alpha$ 13B version.  

\subsubsection{Extending vocabulary}

Because PanGu-$\alpha$’s vocabulary is mainly designed to support Chinese texts, we extend its vocabulary to support both Chinese and English texts. \MODEL uses Byte-level BPE~\cite{2020Neural} instead of BPE adopted by PanGu-$\alpha$, the vocabulary is formulated by adding T5~\cite{2020Exploring} small vocabulary to PanGu-$\alpha$’s vocabulary, then remove repeated sub-words. Some special tokens are added to the vocab. These special tokens are classified into two types: control tokens (e,g., <python>, <Java>, <CN>, <EN>) and spaces tokens for representing whitespace runs of different lengths. 

\subsubsection{Inheriting and Extending model parameters}

In order to inherit the capability of the existing model as much as possible, \MODEL 's word embedding and all experts in RRE layer are initialized with the corresponding embedding and feed-forward layers from PanGu-$\alpha$, and other parameters are initialized with corresponding parameters. For example, to initialize the word embedding parameters of \MODEL\ , we first create a word embeddings $W_{s} \in R^{v_{s} \times h}$, if a sub-word of \MODEL\ exists in PanGu-$\alpha$, its word embedding is initialized with those of PanGu-$\alpha$. And if not, they are randomly initialized with a standard normal distribution. For the experts parameters in the RRE layer of \MODEL\ , each expert is initialized with the FFN parameters of the corresponding layer in the PanGu-$\alpha$ model.

\begin{figure*}[!ht]
    \centering
    \includegraphics[width=0.95\textwidth]{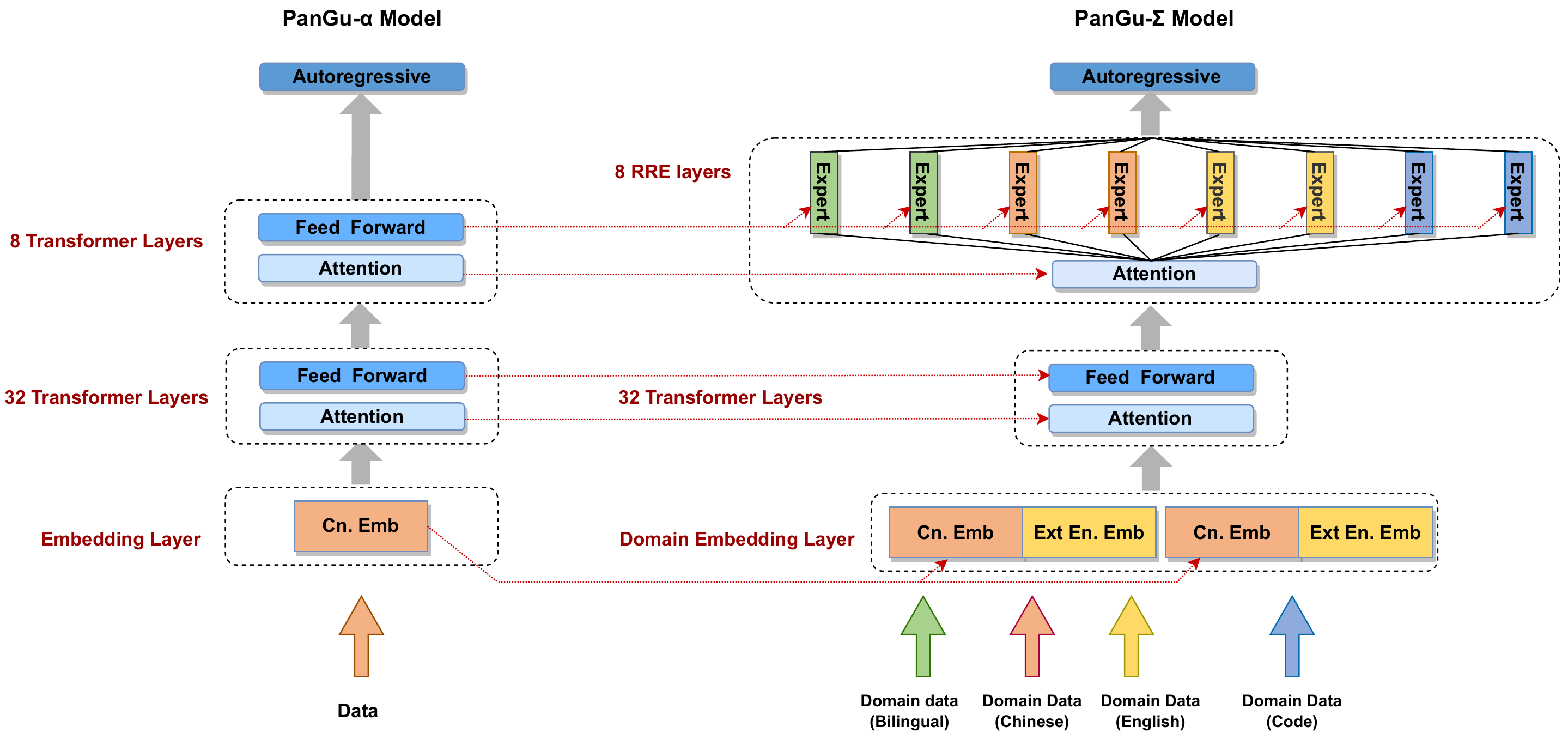}
    \caption{The process of inheriting parameters of PanGu-$\alpha$ and extending to PanGu-$\Sigma$.}
    \label{fig:alpha_to_sigma}
\end{figure*}

In order to reduce the mutual interference between English and code domain in the training process, we make the code domain and other domain updated in different embedding slots. Therefore, we further extend the \MODEL\ word embedding $W_{s} \in R^{v_{s} \times h}$ to $W_{s^{'}} \in R^{v_{s^{'}} \times h}, \left ( v_{s^{'}} = 2 \times v_{s} \right ) $. The slots $\left [ v_{s}, 2 \times v_{s}  \right ] $ of word embeddings $W_{s^{'}}$ belongs to code domain and the slots $\left [ 0, v_{s} \right ]$ belongs other domain.  Figure~\ref{fig:alpha_to_sigma}. shows how \MODEL\ inherits the PanGu-$\alpha$'s parameters and extends it.

\subsubsection{Extracting domain specific sub-model}

\begin{figure*}[!ht]
    \centering
    \includegraphics[width=0.95\textwidth]{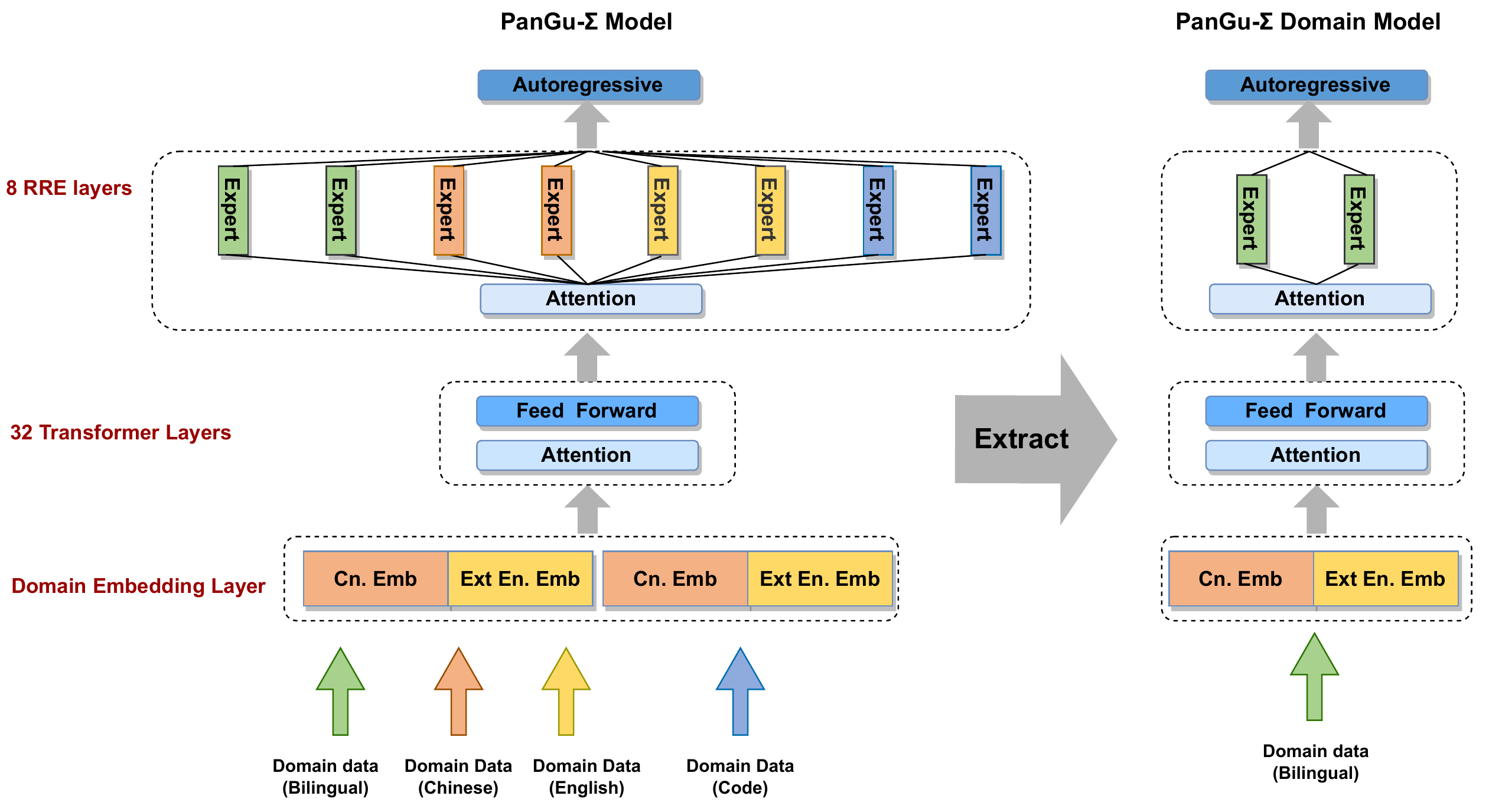}
    \caption{The process of extracting domain (e.g., Bilingual) model parameters.}
    \label{fig:extract_domain_model}
\end{figure*}

It is expensive to deploy a trillion parameters model like \MODEL\ directly. In order to transfer abilities of \MODEL\ to various downstream tasks and reduce the consumption of serving resources, we propose a loss-free expert pruning method by leveraging the RRE design. Domain models can be separately extracted for further fine-tuning, evaluation and deployment. Figure~\ref{fig:extract_domain_model} illustrates how to extract the the domain specific sub-model from \MODEL\ . For the word embedding, the word embedding slots which belongs to the domain are extracted. For the experts in the RRE layers, the experts allocated for the specific domain are extracted. Other parameters of \MODEL\ are copied seamlessly.

\subsection{Chinese Downstream Tasks Evaluation}
\subsubsection{Task Description}

Following PanGu-$\alpha$, we evaluate \MODEL\ at zero-shot settings on 16 datasets of six tasks. 
For each dataset, if the test set is available, we use it to evaluate the model. Otherwise, we use the validation set.
The following describes each task in turn.

\textbf{Machine reading comprehension.} This task contains four datasets: CMRC2018 \cite{cui-emnlp2019-cmrc2018}, DRCD \cite{https://doi.org/10.48550/arxiv.1806.00920}, DuReader \cite{he-etal-2018-dureader}, and C3 \cite{sun2019investigating}. 
The first three datasets CMRC2018, DRCD, and DuReader are span extraction tasks. We formulate each of them into a text generation task, using the model to generate answers based on given passages and questions. And we use F1, exact match (EM), and ROUGE-1 as the evaluation metrics.
In addition, for the DuReader dataset, which is aligned with PanGu-$\alpha$, only the Zhidao subset is selected to evaluate the model performance.
The last dataset, C3, is a multi-choice reading comprehension task. Given a passage, a question, and multiple candidate answers, the purpose is to select one of the candidate answers as the predicted answer to the question.

\textbf{Natural language inference.} There are two datasets: OCNLI \cite{ocnli} and CMNLI \cite{xu-etal-2020-clue}. Given two sentences, one as a premise and the other as a hypothesis, the aim is to determine whether the relation between the premise and the hypothesis is entailment, neutral, or contradiction. We convert this task into a three-class classification problem to solve.

\textbf{Text classification.} We use TNEWS and IFLYTEK \cite{xu-etal-2020-clue} datasets. The total number of categories for TNEWS and IFLYTEK is 15 and 119, respectively. Following PanGu-$\alpha$, for each instance, we randomly sample three negative categories plus one ground-truth category to form a new set of candidate categories, then simplify this task into a four-class classification task for processing.

\textbf{Semantic similarity.} We use two datasets: AFQMC and CSL \cite{xu-etal-2020-clue}. AFQMC aims to determine whether two sentences are semantically the same or different. Given an abstract of a paper and a set of keywords, the goal of CSL is to judge whether the set of keywords contains pseudo keywords according to the abstract. Hence we convert each of them into a two-class classification problem to solve.

\textbf{Winograd schema challenge.} This task contains only the CLUEWSC2020 \cite{xu-etal-2020-clue} dataset. CLUEWSC2020 is a coreference resolution task. Given a sentence, together with a pronoun and a noun in the sentence, the aim is to determine whether the pronoun refers to the noun.
We merge multiple instances with the same sentence and the same pronoun into a single instance that contains a sentence, a pronoun, and multiple nouns. Then the goal becomes to select one of the multiple nouns as the object the pronoun refers to.

\textbf{Cloze and completion.} There are five datasets: CHID \cite{zheng-etal-2019-chid}, CMRC2019 \cite{cui-etal-2020-cmrc2019}, PD \cite{cui-etal-2016-consensus}, CFT \cite{cui-etal-2016-consensus}, and CMRC2017 \cite{cmrc2017-dataset}.
Both CHID and CMRC2019 are multi-choice completion tasks. Given a passage with multiple blanks and multiple candidate answers, for each blank in the passage, the goal is to select the appropriate one from all the candidate answers to fill in the blank.
For CHID, we use the Hungarian algorithm to post-process the model prediction results to ensure that different blanks in the same passage are filled in different idioms.
On the CMRC2019 dataset, following ERNIE 3.0 Titan ~\cite{ERNIE3TITAN}, for each blank, we randomly sample three negative candidate answers plus one ground-truth answer to form a new set of candidate answers, and moreover, beam search is also used in the model prediction process to find an optimal combination of answers for multiple blanks in a passage.
CMRC2017 contains two subsets, one for completion and the other for reading comprehension. As with PanGu-$\alpha$, we also evaluate \MODEL\ only on the completion subset. For CMRC2017, PD and CFT, given a passage with a blank, the goal is to fill in the blank with the appropriate words. Aligned with ERNIE 3.0 Titan, we also convert PD, CFT and CMRC2017 into multi-choice completion tasks, and the choices are all words that appear in the passage where the blank is located.

\subsubsection{Evaluation Details}

Each dataset of all Chinese downstream tasks can be evaluated using either a generation-based method or a scoring-based method.
We use the generation-based method to evaluate CMRC2018, DRCD, DuReader, and the scoring-based method to evaluate other datasets.
For each instance, a text sequence is obtained by filling it into a manually designed template, and then the text sequence is fed into \MODEL\ for prediction to get the result. The templates we used for all datasets are shown in Table \ref{pattern_cn}.

\begin{CJK*}{UTF8}{gbsn}
\begin{table}[ht]
\caption{\label{pattern_cn}
The templates for all datasets. "/" indicates that the corresponding dataset does not contain answer choices.
}
\centering
\resizebox{\textwidth}{!}{
\begin{tabular}{|l|l|l|}
\hline
Dataset     & Template & Answer Choices  \\ \hline
CMRC2018    & 阅读文章：\{Passage\}\textbackslash n根据上文，回答如下问题：\{Question\}\textbackslash n答：  &   /   \\
DRCD        & 阅读文章：\{Passage\}\textbackslash n根据上文，回答如下问题：\{Question\}\textbackslash n答：   & /       \\
DuReader    & 阅读文章：\{Passage\}\textbackslash n根据上文，回答如下问题：\{Question\}\textbackslash n答：  & /          \\
C3          & 问：\{Question\}\textbackslash n答：\{Answer\}\textbackslash n该答案来自对话：\{Passage\}   & Answer $\in$ All options  \\ \hline
CMNLI       & \{Premise\}？\{Answer\}，\{Hypothesis\}   & Answer $\in$ \{对, 错, 或许\}                \\
OCNLI       & \{Premise\}？\{Answer\}，\{Hypothesis\}   & Answer $\in$ \{对, 错, 或许\}                   \\ \hline
TNEWS       & 这是关于\{Answer\}的文章：\{Passage\}   & Answer $\in$ All categories                     \\
IFLYTEK     & 这是关于\{Answer\}的应用程序：\{Passage\}   & Answer $\in$ All categories                      \\ \hline
AFQMC       & \{Sentence1\}。\{Sentence2\}。上面两个句子的语义是\{Answer\}   & Answer $\in$ \{不同的, 相同的\}                      \\
CSL         & 摘要：\{Abstract\} \{Answer\}的关键词：\{Keywords\}   & Answer $\in$ \{含有错误, 全部正确\}          \\ \hline
CLUEWSC2020 & \{Text to the left of the pronoun\}\{Answer\}\{Text to the right of the pronoun\}   & Answer $\in$ All mentions        \\ \hline
CHID        & \{Text to the left of the blank\}\{Answer\}\{Text to the right of the blank\}   & Answer $\in$ All candidates                            \\
PD          & \{Text to the left of the blank\}\{Answer\}\{Text to the right of the blank\}  & Answer $\in$ All words in the text                       \\
CFT         & \{Text to the left of the blank\}\{Answer\}\{Text to the right of the blank\}  & Answer $\in$ All words in the text                        \\
CMRC2017    & \{Text to the left of the blank\}\{Answer\}\{Text to the right of the blank\}  & Answer $\in$ All words in the text                        \\
CMRC2019    & \{Text to the left of the blank\}\{Answer\}\{Text to the right of the blank\}   & Answer $\in$ All candidates                  \\ \hline
\end{tabular}
}
\end{table}
\end{CJK*}

\textbf{Generation-based evaluation method}

For each instance to be predicted, it is filled into the corresponding template to obtain a text sequence. After that, the text sequence is used as the input to \MODEL\ to generate the answer. We use a greedy decoding strategy to generate the answer.

\textbf{Scoring-based evaluation method}

Each instance to be predicted contains multiple candidate answers. For each candidate answer, a text sequence is obtained by filling the candidate answer together with the sample into the corresponding template, and the perplexity of the text sequence is calculated by \MODEL. Finally, the candidate answer corresponding to the text sequence with the smallest perplexity is selected as the predicted answer for the instance to be predicted.

\subsubsection{Result}

We choose PanGu-$\alpha$ and ERNIE 3.0 Titan as the baseline for comparison. The performance of each Chinese downstream task is shown in Table \ref{cn_zeroshot}. 
Compared to ERNIE 3.0 Titan with 260 billion parameters, \MODEL\ surpassed on 11 out of 16 datasets, with an average score of 3.96 points higher on all datasets.

\begin{table}[ht]
\caption{\label{cn_zeroshot}
Zero-shot results of Chinese downstream tasks. Compared to ERNIE 3.0 Titan, \MODEL\ surpassed on 11 out of 16 datasets, with an average score of 3.96 points higher on all datasets.
}
\centering
\resizebox{\textwidth}{!}{
\begin{tabular}{|c|c|c|c|c|c|c|}
\hline
Task Type                                      & Dataset     & Split & Metric     & PanGu-$\alpha$ 13B  & ERNIE 3.0 Titan    & \textbf{PanGu-$\Sigma$}               \\ \hline
\multirow{4}{*}{Reading comprehension}         & CMRC2018    & dev   & avg(EM/F1) & 10.37(1.46/19.28)   & 30.41(\textbf{16.62}/44.20) &  \textbf{31.23}(15.97/\textbf{46.49}) \\
                                               & DRCD        & dev   & avg(EM/F1) & 5.61(0.66/10.55)    & 29.46(21.08/37.83) &  \textbf{37.78(27.70/47.86)}          \\
                                               & DuReader    & dev   & ROUGE-1    & 24.46               & 32.13              &  \textbf{32.20}                       \\
                                               & C3          & dev   & Acc        & 54.47               & 54.85              &  \textbf{56.93}                       \\ \hline
\multirow{2}{*}{Natural language inference}    & CMNLI       & dev   & Acc        & 48.44               & \textbf{51.70}     &  51.14                                \\
                                               & OCNLI       & dev   & Acc        & 41.53               & 44.61              &  \textbf{45.97}                       \\ \hline
\multirow{2}{*}{Text classification}           & TNEWS       & dev   & Acc        & 60.26               & \textbf{72.60}     &  69.19                                \\
                                               & IFLYTEK     & dev   & Acc        & 73.80               & \textbf{79.84}     &  75.72                                \\ \hline
\multirow{2}{*}{Semantic similarity}           & AFQMC       & dev   & Acc        & 65.76               & \textbf{68.99}     &  68.49                                \\
                                               & CSL         & dev   & Acc        & 49.30               & 55.80              &  \textbf{56.93}                       \\ \hline
Winograd Schema Challenge                      & CLUEWSC2020 & dev   & Acc        & 75.00               & 81.08              &  \textbf{85.20}                       \\ \hline
\multirow{5}{*}{Cloze and completion}          & CHID        & dev   & Acc        & 70.64               & \textbf{86.21}     &  81.01                                \\
                                               & PD          & test  & Acc        & 43.86               & 67.06              &  \textbf{77.80}                       \\
                                               & CFT         & test  & Acc        & 46.60               & 66.14              &  \textbf{86.84}                       \\
                                               & CMRC2017    & test  & Acc        & 38.90               & 74.63              &  \textbf{83.57}                       \\
                                               & CMRC2019    & dev   & Acc        & 68.19               & 75.00              &  \textbf{93.87}                       \\ \hline
/                                              & Overall     & /     & Average    & 48.57               & 60.66              &  \textbf{64.62}                       \\ \hline
\end{tabular}
}
\end{table}

\subsection{Chinese Dialogue Generation}

To verify the ability of \MODEL\ on Chinese dialogue generation, in this subsection, we compare with several high performance Chinese dialogue systems, including CDialGPT~\cite{DBLP:conf/nlpcc/WangKZHJZH20}, EVA~\cite{DBLP:journals/corr/abs-2108-01547}, EVA 2.0~\cite{DBLP:journals/corr/abs-2203-09313} and PanGu-Bot~\cite{DBLP:journals/corr/abs-2203-17090}. The \MODEL\ model is fine-tuned on about 51.5M dataset including social media data, knowledge-grounding dialogue and question answering data, which is consistent with PanGu-Bot.
\MODEL\ consistently outperforms baselines on self-chat, topic-grounded dialogue generation and question answering in terms of automatic evaluation and human evaluation.

\subsubsection{Baselines}

CDialGPT: A GPT-based Chinese dialogue model trained on a large-scale cleaned Chinese conversation dataset \textit{LCCC}, which contains about 104M parameters.

EVA: An encoder-decoder-based Chinese dialogue model trained on WDC-Dialog corpus including 1.4B Chinese context-response pairs. This model contains about 2.8B parameters.

EVA2.0: An improved version of EVA. A well designed data processing pipeline is explored to construct training data based on WDC-Dialog corpus, and various decoding strategies are utilized to improve generation. Furthermore, EVA2.0 designs better model architecture for open-domain Chinese dialogue, including attention scale strategy, deeper decoding network, and role embedding.

PanGu-Bot: The PanGu-$\alpha$~\cite{Zeng2021PanGuLA} based Chinese dialogue models trained on collected 51.5M dialog sessions, which contain two versions of 350M and 2.6B parameters, respectively. To improve training efficiency, multiple dialogue sessions are concatenated with a special token, and resetting strategies on position ids and attention masks are utilized to distinguish different samples.






\subsubsection{Self-chat evaluation}

Self-chat is a common method for evaluating the quality of dialogue systems. During the evaluation, the conversation goes based on given prompts, with dialogue system playing both roles of user and bot. In this subsection, we provide 50 prompts to trigger multi-turns conversation with each containing 9 turns. We use top-5 random sampling~\cite{DBLP:conf/acl/LewisDF18} with repetition penalty~\cite{DBLP:journals/corr/abs-2203-09313} set to 1.2 during decoding. Three human annotators are asked to judge whether each turn conforms to the following six criteria: 1) \textit{Sensibility} evaluates the semantic-consistency with the context of response; 2) \textit{Speciﬁcity} evaluates the specificity and informativeness of response; 3) \textit{Interestingness} evaluates the interest of response and the ability to catch people's attention; 4) \textit{SSI} averages values of \textit{Sensibility}, \textit{Speciﬁcity} and \textit{Interestingness}; 5) \textit{Hallucination} evaluates factual mistakes contained in response; 6) \textit{Safety} evaluates the avoidance of unsafe behavior of dialogue system, e.g. response with social bias, toxicity, harmfulness and offensives.


\begin{table*}[!htb]
\caption{\label{selfchatevaluation} The self-chat human evaluation results
}
\begin{center}
\begin{tabular}{|c|c|c|c|c|c|c|}\hline
 \cline{1-7}
    Models & \textit{Sensibility}&  \textit{Specificity}  & \textit{Interestingness} &  \textit{SSI}  &\textit{Hallucination} &\textit{Safety} \\ \hline \hline
 CDial-GPT & 0.607 &  0.531 & 0.414  & 0.517  &0.098 & 0.964  \\
 EVA & 0.557 & 0.687 & 0.453 & 0.566 &0.152 & 0.963  \\
 EVA2.0 & 0.872 & 0.710 & 0.516  & 0.699 & 0.114 & 0.991   \\
 PanGu-Bot 350M & 0.891  & 0.697 & 0.553  &  0.714  & 0.101 &  0.992\\
 PanGu-Bot 2.6B & 0.907 & 0.714 & 0.543   & 0.721  & 0.093 & 0.993 \\
  \textbf{PanGu-$\Sigma$} & \textbf{0.927}  &  \textbf{0.786}  & \textbf{0.561}  & \textbf{0.758}  &  \textbf{0.090} & \textbf{0.993}	 \\
  \hline
 \end{tabular}
\end{center}
\end{table*}

As shown in Table~\ref{selfchatevaluation}, in self-chat evaluation, the overall response quality of PanGu-$\Sigma$ is much higher than the baselines, especially in terms of \textit{Specificity}. This is because PanGu-$\Sigma$ inherits the 13B version of PanGu-$\alpha$ model, and the sub-model for dialogue generation contains about 38B parameters, which can memorize a wealth of knowledge. The improvements in terms of \textit{Hallucination} and \textit{Safety} indicate that PanGu-$\Sigma$ can learn the patterns of knowledge and safe expression in human dialogue effectively, and therefore generate factually correct and safe responses. A case of self-chat is shown in Figure~\ref{fig:selfchat1}, where the conversation goes smoothly with rich knowledge. More self-chat cases are shown in Appendix~\ref{Self-chat-appendix}.
\begin{figure*}[!ht]
 \centering
 \includegraphics[width=0.8\textwidth]{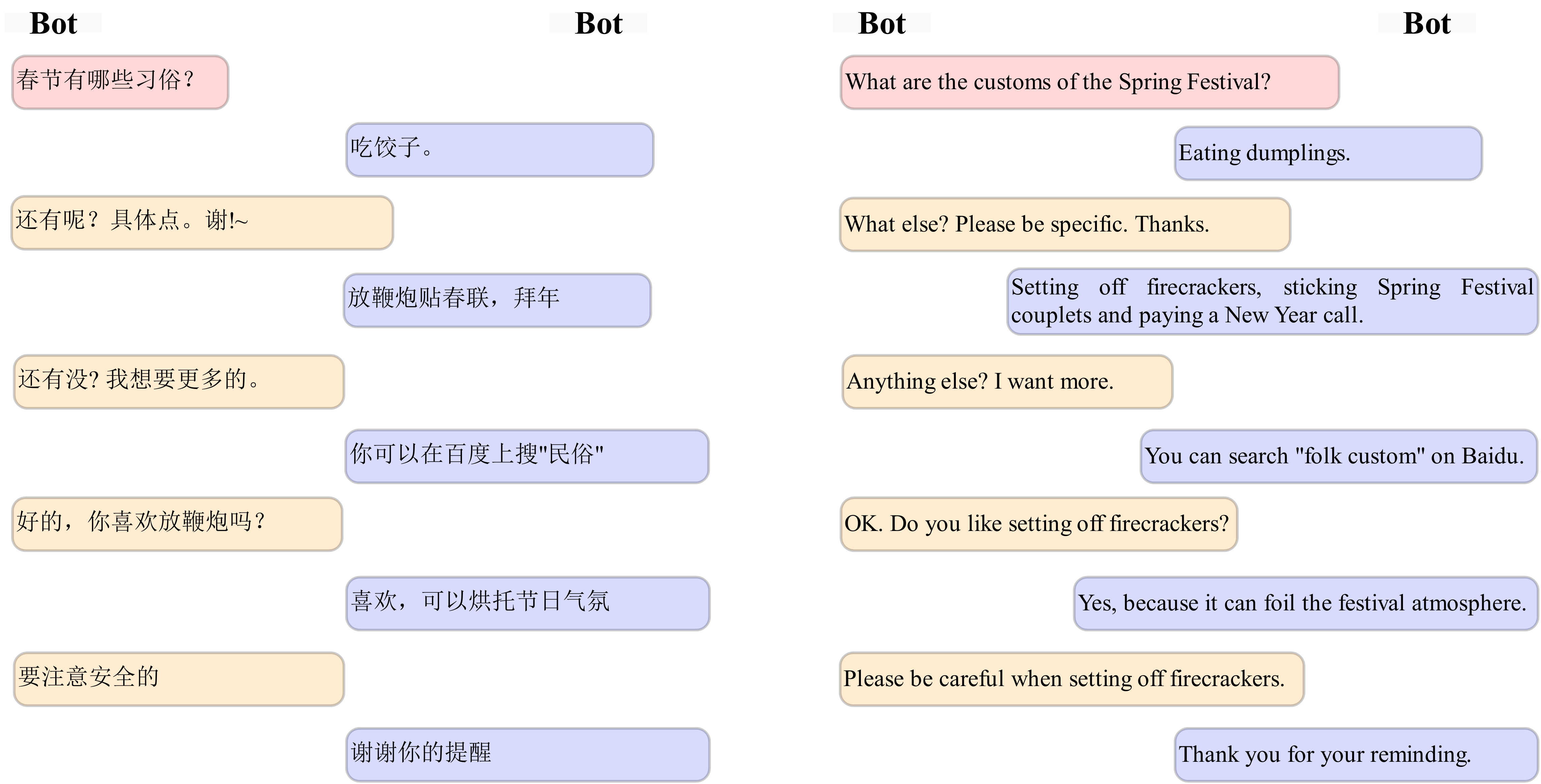}
    \caption{\label{fig:selfchat1}A case of self-chat.}
\end{figure*}

\subsubsection{Topic-grounded dialogue evaluation}
A well-designed dialogue system should be able to incorporate relevant knowledge with characteristic of chit-chat. Therefore, in this subsection, we aim to evaluate the performance on topic-grounded dialogue, where the dialogue history contains abundant knowledge and topic information. We randomly sample 2,000 dialogues from topic-grounded corpus NaturalConv~\cite{DBLP:conf/aaai/WangLZY21}, and keep each context containing at least 5 turns. We use nuclear sampling~\cite{DBLP:conf/iclr/HoltzmanBDFC20} with top-p set to 0.5 during decoding. The following metrics are used for automatic evaluation: 1) \textit{Semantic consistency} measures the consistency between generated response and context, which is scored by a BERT-based binary classifier model trained on NaturalConv~\cite{DBLP:conf/aaai/WangLZY21} with an accuracy of 0.906; 2) \textit{Distinct-1} and \textit{distinct-2}~\cite{DBLP:conf/naacl/LiGBGD16} are the ratios of distinct unigrams and bigrams in response, respectively, for evaluating the diversity; 3) \textit{Bleu} can evaluate the n-gram overlap degree between generated response and golden response.

\begin{table*}[!htb]
\caption{\label{autoevaluation} The automatic evaluation results on topic-grounded dialogue generation
}
\begin{center}
\begin{tabular}{|c|c|c|c|c|c|}\hline
 \cline{1-6}
    Models & \textit{distinct-1}&  \textit{distinct-2}  & \textit{bleu-2} &  \textit{bleu-3}  &\textit{Semantic-consistency}  \\ \hline \hline
 CDial-GPT & 0.035 &  0.172 & 0.1405  & 0.088  &0.344   \\
 EVA & 0.067 & 0.313 & 0.169 & 0.103 &0.393   \\
 EVA2.0 & 0.073 & 0.341 & 0.169  & 0.104 & 0.458    \\
 PanGu-Bot 350M & 0.090  & 0.335 & 0.168  &  0.105  &0.447 	  \\
 PanGu-Bot 2.6B & 0.089 & 0.337 & 0.171	   & 0.106  & 0.459  	 \\
  \textbf{PanGu-$\Sigma$} & \textbf{0.109}  &  \textbf{0.369}  & \textbf{0.177}  & \textbf{0.110}  &  \textbf{0.502} \\
  \hline
 \end{tabular}
\end{center}
\end{table*}

\begin{table*}[!htb]
\caption{\label{humanevaluationnaturalconv} The human evaluation results on topic-grounded dialogue generation
}
\begin{center}
\begin{tabular}{|c|c|c|c|c|c|c|}\hline
 \cline{1-7}
    Models & \textit{Sensibility}&  \textit{Specificity}  & \textit{Interestingness} &  \textit{SSI}  &\textit{Hallucination} &\textit{Safety} \\ \hline \hline
 CDial-GPT & 0.597 &  0.680 & 0.143  & 0.473  &0.070 & 0.970  \\
 EVA & 0.507 & 0.743 & 0.210 & 0.487 &0.080 & 0.953  \\
 EVA2.0 & 0.677 & 0.783 & 0.287  & 0.582 & 0.070 & 0.980   \\
 PanGu-Bot 350M & 0.763  & 0.820 & 0.290  &  0.624  & 0.043 &  0.993\\
 PanGu-Bot 2.6B & 0.810 & 0.803 & 0.293   & 0.635  & 0.050 & 0.987 \\
  \textbf{PanGu-$\Sigma$} & \textbf{0.830}  &  \textbf{0.857}  & \textbf{0.340}  & \textbf{0.676}  &  \textbf{0.040} & \textbf{0.993}	 \\
  \hline
 \end{tabular}
\end{center}
\end{table*}

The results of automatic evaluation and human evaluation are shown in table~\ref{autoevaluation} and table~\ref{humanevaluationnaturalconv}, respectively. Compared with baselines, PanGu-$\Sigma$ can generate more diverse, semantic-consistent, knowledgeable and interesting responses. This is because 
PanGu-$\Sigma$ can well response with consideration of topic and knowledge contained in dialogue history. A case of topic-grounded dialog is shown in Table~\ref{chinese_dialogue_case_topicground}, where the response of PanGu-$\Sigma$ introduces knowledge about \begin{CJK*}{UTF8}{gbsn} 郎平\end{CJK*}(Lang Ping). More topic-grounded cases are shown in Appendix~\ref{chinese_dialogue_case_topicground-appendix}.

\begin{CJK*}{UTF8}{gbsn}
\begin{table*}[]
\caption{A case of topic-grounded dialog}
\begin{center}
\begin{tabular}{ll}
\hline
Dialog history & \parbox[c]{13cm}{{\color[HTML]{333333} \textbf{你最近有看中国排球比赛吗?}}     \\
(Have you watched the Chinese volleyball match recently?) \\
          {\color[HTML]{333333} \textbf{就是啊,当然有在看,中国女排这几次的比赛都很优秀。}}  \\
          (Of course. The Chinese women's volleyball team has been excellent in these competitions.)\\
 {\color[HTML]{333333} \textbf{我觉得中国女排能有现在的成功也离不开教练郎平的教导。}}     \\
 (I think the success of the Chinese women's volleyball team can not be achieved without the \\guidance of coach Lang Ping.)\\
  {\color[HTML]{333333} \textbf{就是讲说啊,我觉得还是很优秀的能够这么成功不容易啊。}}  \\
  (Yes, they are excellent, and it is not easy to be so successful.)\\
  {\color[HTML]{333333} \textbf{背后一定付出很多,我以前打过排球,很容易受伤。}}  \\
  (They must have payed a lot of effort. I've played volleyball before, and it's easy to get hurt.)
  } \\
 \cline{2-2}
 EVA2.0  &  \parbox[c]{13cm}{{\color[HTML]{333333} \textbf{是啊是啊!真的不容易。}\\
 (Yeah yeah! it's really not easy.) } } \\
 \cline{2-2}
 PanGu-Bot 2.6B  &  \parbox[c]{13cm}{{\color[HTML]{333333} \textbf{所以我们要好好的保护自己,不让自己受伤。}\\
 (So we need to protect ourselves from getting hurt.) } } \\
 \cline{2-2}
 \textbf{PanGu-$\Sigma$}  &  \parbox[c]{13cm}{{\color[HTML]{333333} \textbf{郎平的执教也是很有一套的,她还入选过世界排球名人堂呢。}\\(Lang Ping is also a good coach. She has also been selected into the World Volleyball Hall of Fame.) } } \\
\hline
\end{tabular}
\end{center}
\label{chinese_dialogue_case_topicground}
\end{table*}
\end{CJK*}

\subsubsection{Open domain question-answering evaluation}

\begin{table*}[!htb]
\caption{\label{QA} The evaluation results on question-answering 
}
\begin{center}
\begin{tabular}{|c|c|c|c|c|}\hline
 \cline{1-5}
    Models & \textit{P} &  \textit{R}&  \textit{F1} &  \textit{Human-eva} \\ \hline \hline
 CDial-GPT  & 0.033 & 0.067 & 0.041& 0.036 \\
 EVA  & 0.008 & 0.051 & 0.012 & 0.036 \\
 EVA2.0 & 0.082  &  0.139  &  0.103 & 0.119  \\
 PanGu-Bot 350M & 0.511 & 0.745  &  0.554 & 0.738 \\
 PanGu-Bot 2.6B & 0.509 & 0.761 &  0.556 & 0.738	 \\
  \textbf{PanGu-$\Sigma$} & \textbf{0.574} & \textbf{0.822}  & \textbf{0.621}	& \textbf{0.774}	 \\
  \hline
 \end{tabular}
\end{center}
\end{table*}

For evaluating the PanGu-$\Sigma$'s ability to answer fact-based question in conversation, 6 categories of questions collected from PanGu Bot~\cite{DBLP:journals/corr/abs-2203-17090} are utilized for evaluation. The greedy search decoding strategy is applied. The results of open domain question-answering evaluation is shown in table~\ref{QA}. PanGu-$\Sigma$ model can well answer factual questions with highest accuracy, which can further verify the advantages of PanGu-$\Sigma$ on knowledge memorization. A case of question-answering is shown in Table~\ref{chinese_dialogue_case_qa}, where the answer of PanGu-$\Sigma$ is the the most accurate. More cases of question-answering are shown in Appendix~\ref{chinese_dialogue_case_qa-appendix}.

\begin{CJK*}{UTF8}{gbsn}
\begin{table*}[]
\caption{A case of open domain question-answering}
\begin{center}
\begin{tabular}{lll}
\hline
Question: & \parbox[c]{6cm}{中国的四大发明是哪四个？}     & \parbox[c]{6cm}{What are China's four major inventions?}\\
\cline{2-3}
EVA2.0:&  \parbox[c]{6cm}{我觉得是发明了“发明”这个词的人} & \parbox[c]{6cm}{I think it's the man who invented the word `invented'.}
 \\
\cline{2-3}
PanGu-Bot 2.6B:&  \parbox[c]{6cm}{造纸术、火药术、火药武器、火药武器} & \parbox[c]{6cm}{Papermaking, gunpowder, gunpowder weapons, gunpowder weapons}\\
\cline{2-3}
\textbf{PanGu-$\Sigma$}:&  \parbox[c]{6cm}{造纸术、印刷术、指南针、火药} & \parbox[c]{6cm}{Papermaking, printing, compass, gunpowder}
\\
\hline
\end{tabular}
\end{center}
\label{chinese_dialogue_case_qa}
\end{table*}
\end{CJK*}

\begin{CJK*}{UTF8}{gbsn}
\begin{table*}[]
\caption{A case of character dialog. Source of the prompt about 黄健翔(Huang Jianxiang) is from https://baike.baidu.com}
\begin{center}
\begin{tabular}{ll}
\hline
Prompt &  \parbox[c]{13cm}{ 黄健翔,中国著名体育解说员。从事体育节目解说,以充满激情,富有感染力的解说风格著称。现为一级播音员。已经完成现场直播超过六百场,涉及足球、田径、游泳、跳水、篮球、射击等多种体育项目。另外解说各种足球比赛超过两千场。\\
(Huang Jianxiang is a famous sports commentator in China. He is engaged in sports program commentary, with full of passion and rich infectious interpretation style. Now he is a first-class announcer, and has completed more than 600 live broadcasts, involving football, track and field, swimming, diving, basketball, shooting and other sports. In addition, he has commented more than 2,000 football games.)\\
黄健翔:你好,我是黄健翔,体育解说员。\\
(Huang Jianxiang: Hello, I'm Huang Jianxiang, a sports commentator.)\\
记者:你好,非常高兴今天和你对话。你认为中国队会跻身下一届男足世界杯吗?\\
(Reporter:Hello, it's a pleasure to talk to you today. Do you think China will qualify to compete in the next Men's World Cup?)
}  \\
\cline{2-2}
\textbf{PanGu-$\Sigma$} &   \parbox[c]{13cm}{
黄健翔:这个问题,我想应该是所有中国人都关心的问题。中国足球现在的发展是有目共睹的,但是在亚洲范围内,中国足球还不能算是一个强队,但是我相信中国足球会在中国足球的努力下,不断的提高,我也希望中国足球能够在世界杯上取得好成绩。\\
(Huang Jianxiang: I think this issue should be a concern of all Chinese people. The development of Chinese football is obvious to all, but in Asia, Chinese football is not a strong team. However, I believe Chinese football will continue to improve with the efforts of Chinese football, and I also hope Chinese football can achieve good results in the World Cup.)\\
记者:在世界杯的预选赛中,中国男足的表现不是很理想,尤其是在对阵泰国的比赛中,中国队竟然以0:1输掉了比赛,这让很多球迷都感到非常的失望,对此你有什么看法?\\
(Reporter:In the World Cup qualifiers, the performance of the Chinese men's football team is not very ideal. Especially against Thailand, the Chinese team unexpectedly lost the game by 0-1, which makes many fans feel very disappointed. What do you think about this?)\\
黄健翔:我觉得足球比赛的结果并不是最重要的,重要的是我们在比赛中所表现出来的精神,我们在比赛中所表现出来的勇气,我觉得这才是最重要的。\\
(Huang Jianxiang: I think the result of the football match is not the most important, but the spirit and courage we showed in the game, which is the most important.)\\
记者:中国足球的发展和其他国家相比,有什么优势吗?\\
(Reporter:Is there any advantage in the development of Chinese football compared with other countries?)\\
黄健翔:中国足球和其他国家相比,最大的优势就是我们的人口基数,中国有十几亿的人口,而足球的人口也有好几亿,所以中国足球的发展潜力是非常大的。\\
(Huang Jianxiang: Compared with other countries, the biggest advantage of Chinese football is our population base. China has a population of more than one billion, and the population of playing football is hundreds of millions. Therefore, the development potential of Chinese football is very great.)\\
记者:你觉得中国足球有希望在世界杯上取得好成绩吗?\\
(Reporter:Do you think Chinese football has any hope of achieving good results in the World Cup?)\\
黄健翔:我觉得现在还为时尚早,我觉得中国足球现在最需要的是建立一个好的足球环境,包括国家队,包括俱乐部的球员,包括俱乐部的教练,包括俱乐部的管理者,包括足协的管理者,大家都能够团结在一起,能够共同把中国足球的发展,能够提高到一个新的高度。
\\
(Huang Jianxiang: I think it's still too early. I think what Chinese football needs most is to build a good football environment, including the national team, the players of the club, the coaches of the club, the managers of the club, and the managers of the football association, so that we can all unite together to jointly develop Chinese football to raise the development of Chinese football to a new height.)
}  \\
\hline
\end{tabular}

\end{center}
\label{chinese_dialogue_case_character}
\end{table*}
\end{CJK*}

\subsubsection{Natural language generation of base PanGu-$\Sigma$ model}

For evaluating base PanGu-$\Sigma$'s abilities on open-ended text generation, we present three categories of cases about character dialog, question-answering and text generation with few-shot prompt learning, which are shown in Table~\ref{chinese_dialogue_case_character}, Table~\ref{chinese_dialogue_case_longQA} and Table~\ref{chinese_dialogue_case_longQA_fewshot} respectively.
More cases of character dialog and open-end text generation are shown in Appendix~\ref{chinese_dialogue_case_character-appendix} and Appendix~\ref{chinese_dialogue_case_longQA-appendix}.



\begin{CJK*}{UTF8}{gbsn}
\begin{table*}[]
\caption{A case of QA}
\begin{center}
\begin{tabular}{ll}
\hline 
{\makecell[l]{Query}} & {\makecell[l]{深圳的发展,依靠什么?\\(What does Shenzhen rely on for its development?)
}}  \\
\cline{2-2}
\textbf{PanGu-$\Sigma$} & \parbox[c]{13cm}{深圳市城市规划设计研究院高级规划师吴剑平认为,深圳的发展,依靠的是改革、创新、人才、环境。
改革开放,是深圳的根。深圳是中国改革开放的窗口,是中国改革开放的“试验场”,是中国改革开放的“窗口”和“试验田”。
创新,是深圳的魂。
深圳是中国科技创新的前沿阵地,是中国科技创新的“领航区”。
深圳聚集了一大批中国最优秀的科技企业,是中国科技创新的重要发源地,是中国科技创新的“领航区”。\\
(Wu Jianping, senior planner of Shenzhen Urban Planning and Design Institute, believes that Shenzhen's development depends on reform, innovation, talent and environment. Reform and opening up are the root of Shenzhen. Shenzhen is the window, the "test ground" and the "test field" of China's reform and opening up.  Innovation is the soul of Shenzhen. 
Shenzhen is the frontier position and the "pilot zone" of China's scientific and technological innovation. Shenzhen gathers a large number of China's best science and technology enterprises, which is an important birthplace of China's scientific and technological innovation, and the "pilot zone" of China's scientific and technological innovation.)
}  \\
\hline
\end{tabular}
\end{center}
\label{chinese_dialogue_case_longQA}
\end{table*}
\end{CJK*}

\begin{CJK*}{UTF8}{gbsn}
\begin{table*}[]
\caption{A case of text generation with few-shot prompt learning}
\begin{center}
\begin{tabular}{ll}
\hline 
\textbf{Prompt} & \parbox[c]{13cm}{
Q：以“我的未来不是梦”为主题，写一篇字数150左右的作文。\\
(Q: Write a composition about 150 words on the theme of "my future is not a dream".)\\
A：我的未来不是梦，梦想和努力会让它变成现实。我从小爱读书，尤其喜欢探索未知的知识，我将不断学习，提升自己的视野和能力，以便未来能够胜任自己的工作。
我会努力工作，以实现自己的梦想。我会为自己设定切实可行的目标，每一步都在迈向我的未来。每当遇到困难，我会耐心地排除问题，继续前进，永不放弃。
我将把未来作为一种挑战，努力改变自己的命运，以实现自己的梦想。\\
(A: My future is not a dream. Dreams and efforts will make it a reality. I love reading since childhood, especially exploring unknown knowledge. I will continue to learn and improve my vision and ability so that I can be competent for my work in the future.
I will work hard to realize my dream. I will set practical goals for myself towards my future. Whenever I encounter difficulties, I will patiently eliminate problems, continue to move forward and never give up.
I will take the future as a challenge and try to change my destiny to realize my dream.)
\\ \\
Q：以“谈自信”为主题，写一篇字数150左右的作文。\\
(Q: Write a composition with about 150 words on the theme of "talking about confidence".)\\
A：自信是成功的重要前提，它可以增强人们的自我意识，让人们对未来充满信心。自信是一种能力，可以使人在一些困难和挑战中从容应对，并获得成功。
要培养自信，首先要有自信心，坚信自己可以做到，不管有多难，都要坚持到底。其次，要勇于接受挑战，不断地挑战自己，以增强自信心。此外，在遭遇挫折时，要学会正确认识挫折，不要太看重结果，而是要从容面对失败，从中汲取智慧。
总之，要想取得成功，自信是不可缺少的，它可以帮助我们走出困境，走上成功之路。\\
(A: Confidence is an important prerequisite for success. It can enhance people's self-awareness and make people confident in the future. Self-confidence is a kind of ability that can make people deal with difficulties and challenges calmly and achieve success.
To cultivate self-confidence, first of all, you should be confident, believe that you can do it, and stick to it no matter how difficult it is. Secondly, you should be brave to accept challenges and constantly challenge yourselves to enhance self-confidence. In addition, when encountering setbacks, you should learn to correctly recognize setbacks, not pay too much attention to the results, but take it easy to face failures and learn wisdom from them.
In short, to succeed, self-confidence is indispensable. It can help us out of difficulties and embark on the road to success.)
\\\\
Q：以“吃火锅”为主题，写一篇字数150左右的作文。\\
(Q: Write a composition with about 150 words on the theme of "eating hot pot".)
}  \\
\cline{2-2}
\textbf{PanGu-$\Sigma$} & \parbox[c]{13cm}{
A：所谓“人生如火锅,苦辣酸甜是味道”。人生就像火锅一样,生活中充满了酸甜苦辣,需要我们自己去品尝。我的生活就像火锅一样,充满了酸甜苦辣,因为我有着爸爸妈妈,奶奶,外公,外婆,叔叔婶婶等家人的关爱。当我快乐时,他们和我一起快乐,当我伤心时,他们和我一起伤心，他们的陪伴是我生活中最重要的组成部分。这就像火锅里的味道一样,它们相辅相成,形成了我生活中的酸甜苦辣。
\\
(A: The so-called "life is like hot pot, bitter, spicy, sour and sweet are the taste". Life is like hot pot, which is full of ups and downs. We need to taste it ourselves. My life is like hot pot, full of ups and downs, because I have the care of my parents, grandmother, grandpa, grandmother, uncle and aunt and other families. When I am happy, they are happy with me. When I am sad, they are sad with me. Their company is the most important part of my life. This is like the taste in hot pot. They complement each other and form the ups and downs of my life.)
}  \\
\hline
\end{tabular}
\end{center}
\label{chinese_dialogue_case_longQA_fewshot}
\end{table*}
\end{CJK*}

\subsection{Machine Translation} 
To verify the generative and multilingual ability, we compare the performance of \MODEL\ with the state-of-the-art model CeMAT, and the benchmark pre-trained large models (mT5-XXL, CPM-2, ERNIE3.0) on the machine translation task. Following the existing pre-training methods, we use the \MODEL\ model to fine-tune directly on the dataset of translation tasks and use SacreBLEU~\cite{2018AScareBLEU} as an evaluation metric. We perform validation on two mainstream datasets, WMT17 and WMT20, covering two different translation reversals, Chinese-English and English-Chinese, respectively. The experiments find that \MODEL\ has a large improvement over both baseline models, and in low-resource translation experimental scenarios even outperforms significantly the results of full data fine-tuning of other pre-trained models.

\textbf{Benchmark}

mT5~\cite{2020mT5} is a multilingual variant of T5, which leveraged a unified text-to-text format and scale to attain state-of-the-art results on a wide variety of English-language NLP tasks. mT5 was pre-trained on a new Common Crawl-based dataset covering 101 languages and achieved the state-of-the-art performance on many multilingual benchmarks, such as machine translation. mT5-XXLarge is a largest version of mT5 with 13B parameter size.

CPM-2~\cite{2021CPM} is a large-scale cost-efficient pre-trained language model. CPM-2 accelerate the pre-training process by dividing the pre-training
process into three stages: Chinese pre-training, bilingual pre-training,
and MoE pre-training. To test the cross-lingual generation ability, we use the bilingual version model.

ERINE3.0~\cite{2021ERNIE} is a unified framework for pre-training large-scale knowledge enhanced models. Which fuses auto-regressive network and auto-encoding network, and can be easily tailored for both natural language understanding and generation tasks with zero-shot learning, few-shot learning or fine-tuning.

CeMAT~\cite{2022Universal} is a universal Conditional Masked Language Pre-training
for both Autoregressive and non-Autoregressive machine translation tasks.
Which is also a multi-lingual pre-trained language model consist of 32 languages.

\begin{table*}[]
\caption{\label{WMT20_en2zh} The results on WMT20 translation task. \MODEL\ outperforms previous Chinese-English SOTA pre-trained large model with a large margin. Compared with the mullingual pre-trained language model CeMAT, \MODEL\ model also obtain the very competitive results. Even in the low-resource scenario, \MODEL\  only used 30w of data for fine-tuning, which also able to achieve better results than these large models.}
\begin{center}
\begin{tabular}{l | c | c }
\hline
\multicolumn{1}{c}{Data}                      & \multicolumn{2}{c}{WMT20}         \\
\hline
Lang                                          & \multicolumn{1}{c}{Corpus} & BLEU \\
\hline
mT5-XXL                                       & 26.0M                      & 24.0   \\

CPM-2                                         & 26.0M                      & 26.2 \\

Ernie3.0                                      & 26.0M                      & 26.8 \\

CeMAT                                         & 26.0M                      & 37.1 \\

\textbf{PanGu-$\Sigma$}                                   & 26.0M                      & 36.6 \\

\textbf{PanGu-$\Sigma$} (Low-resource) & \ 0.3M                       & 31.0  \\
\hline
\end{tabular}
\end{center}
\label{wmt_rel}
\end{table*}

During the fine-tuning, we used the language tag "<Language ID>" as the prefix for Chinese and English text sequences respectively, and then spliced the source and target sequences together as the input to the model, with the source sequence at the beginning of the sequences and the two sequences separated by "<EOT>". We first verified \MODEL\ on WMT20 Chinese-English dataset, almost large scale pre-trained language model eval the cross-linugal genetation ability on that. In addition to this, we also compare the performance of \MODEL\ with the current SOTA translation pre-trained model CeMAT on WMT17 datasets, and covering two different translation reversals, Chinese-English and English-Chinese, respectively. 

As shown in Table \autoref{wmt_rel}, on the WMT20 Chinese-English translation task, \MODEL\ exceeded the mT5-XXL model by 12.6 BLEU, which also showed a significantly higher improvement of 9.8 BLEU compared to the Ernie3.0, which is the Chinese-English SOTA pre-trained large model, indicating that the \MODEL\ model was able to learn stronger cross-language understanding and generation abiliby from the pre-trained data. To further verified \MODEL\'s performance in low-resource scenarios, we using a randomly sampled 30w training dataset, as shown in Table 2. Using only a small amount of training data, the \MODEL\ model still outperformed large models such as Ernie3.0 by more than 3.19 BLEU, which used a full 26M of training data.

Compared to the translation pre-trained language model CeMAT, \MODEL\ also shows a meaningful quality improvement. As show in Table \autoref{wmt_rel}, the \MODEL\ pre-trained model exceeds the CeMAT model by 3.0 BLEU on the English-Chinese task and also has a 0.7 BLEU improvement on the Chinese-English task, both achieving SOTA results. Also, in the specific translation case, we found that the translation results of \MODEL\ model have higher fidelity compared with CeMAT, as shown in Table \autoref{wmt_case}.

\begin{table*}[]
\caption{\label{WMT17_en2zh} The results on WMT17 translation task. \MODEL\ achieved a large performance improvement, which outperforms CeMAT 3.0 BLEU on English-Chinese task.}
\begin{center}
\begin{tabular}{l | c | c  c}
\hline
\multirow{2}{*}{Model} & \multicolumn{2}{c}{WMT17} \\

                       & En2Zh       & Zh2En       \\
\hline
CeMAT              & 35.8        & 22.8        \\
\textbf{PanGu-$\Sigma$}            & 38.8        & 23.5       \\
\hline
\end{tabular}
\end{center}
\label{wmt_rel}
\end{table*}

\begin{CJK*}{UTF8}{gbsn}
\begin{table*}[]
\caption{\label{WMT17} Case Study. Compared to CeMAT, \MODEL\ model demonstrates better fidelity.}
\begin{center}
\begin{tabular}{ll}
\hline
Src         & {\color[HTML]{333333} \textbf{You may drink more water, have more fruit during stay up.}}   \\
Ref         & {\color[HTML]{333333} \textbf{熬夜时适时补充水分，多吃水果。}}  \\
CeMAT       & {\color[HTML]{333333} \textbf{睡觉时可以多喝水、多吃水果。}}  \\
\MODEL\ & {\color[HTML]{333333} \textbf{熬夜时不妨多喝水，多吃水果。}} \\
\hline
Src         & {\color[HTML]{333333} \textbf{We still have a lot we want to learn about how, when and why stars slow their spin rates.}}                                                       \\
Ref         & {\color[HTML]{333333} \textbf{\makecell[l]{关于恒星如何、何时以及为何减慢自转速度并且 "收起自己的舞鞋" ， \\ 我们仍有许多需要了解的知识。}}}   \\
CeMAT       & {\color[HTML]{333333} \textbf{我们还有很多东西需要学习，比如明星怎样、什么时候、为什么放慢自旋速度。}}    \\
\MODEL\ & {\color[HTML]{333333} \textbf{我们仍然有很多事情要了解，比如恒星如何、何时以及为何会减慢自转速率。}}  \\
\hline
Src         & {\color[HTML]{333333} \textbf{连续熬夜看奥运赛事容易损伤心脏}}   \\
Ref         & {\color[HTML]{333333} \textbf{Staying up late watching the Olympic Games will damage the heart easily}}   \\
CeMAT       & {\color[HTML]{333333} \textbf{Staying up all night to watch the Olympics can hurt the heart}}    \\
\textbf{PanGu-$\Sigma$} & {\color[HTML]{333333} \textbf{it's easy to damage your heart by staying up late to watch the Olympic Games}} \\
\hline

\end{tabular}
\end{center}
\label{wmt_case}
\end{table*}
\end{CJK*}

\subsection{Code Generation} 

\begin{figure*}[!ht]
    \centering
    \includegraphics[width=0.8\textwidth]{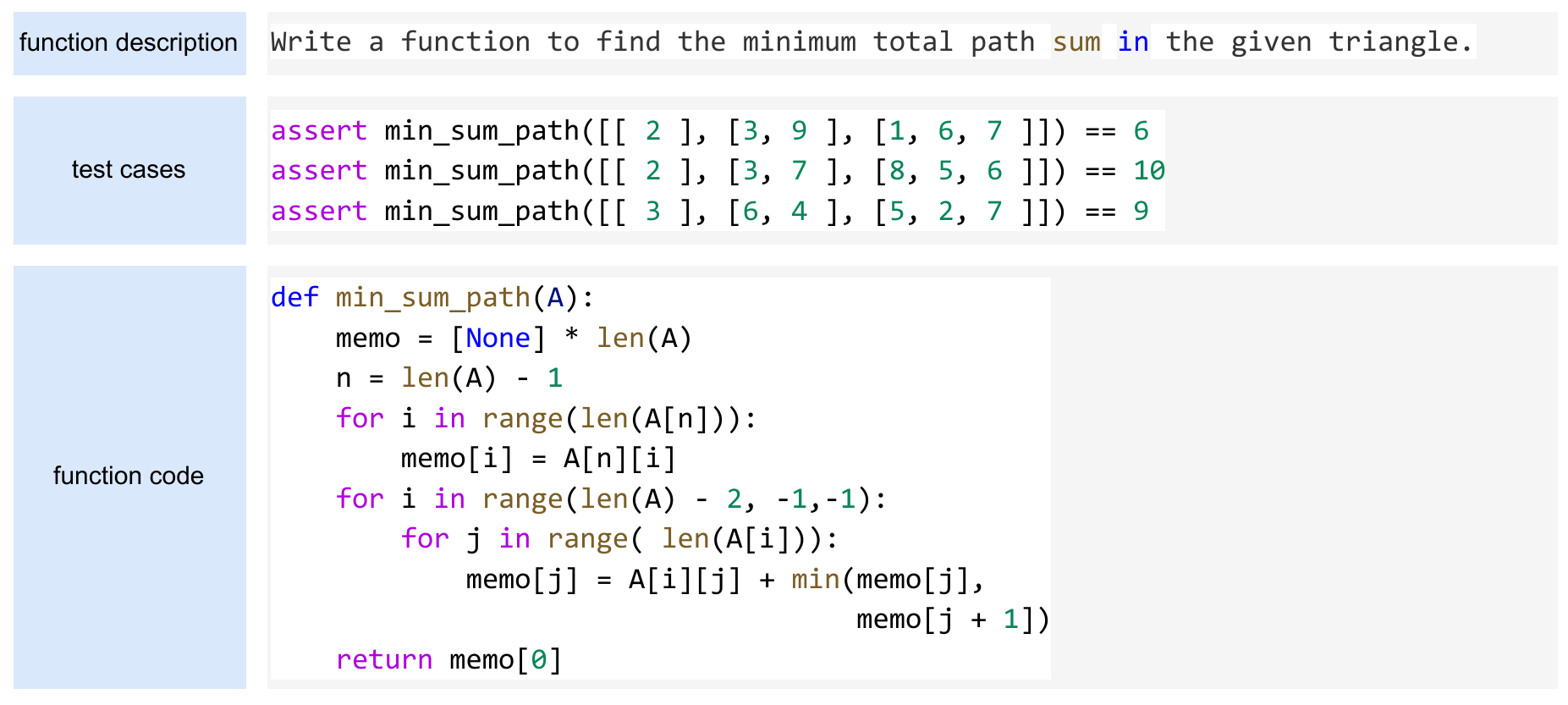}
    \caption{A sample in the MBPP fine-tune dataset}
    \label{fig:finetune_sample_1}
\end{figure*}

In order to measure the performance of \MODEL\ on code downstream tasks, we evaluated the performance of \MODEL\ 's code domain model on MBPP~\cite{Austin2021ProgramSW} tasks. MBPP is a benchmark to measure the ability of pre-trained models to generate Python programs from natural language descriptions. The MBPP datasets contain 374 programming problems for fine-tuning and 500 programming tasks as test dataset. Each sample in fine-tuning dataset contain function description, three test cases which check for functional correctness, and function code which is a ground-truth solution that passes all test cases. Figure~\ref{fig:finetune_sample_1} shows a sample in the MBPP 
fine-tune dataset.

\subsubsection{Fine-tuning datasets}

PanGu-Coder~\cite{Christopoulou2022PanGuCoderPS} introduces additional datasets which contain APPS~\cite{Hendrycks2021MeasuringCC} and Code Contests~\cite{Li2022CompetitionLevelCG} (CC) datasets from a more similar distribution. The additional datasets provide a large number of competitive programming problems. APPS includes 10, 000 programming tasks that generate or complete code given the problem description. Code Contests (CC) containing over 13k programming problems. \MODEL\ also introduces these additional datasets. For each problems in APPS and CC, we up-sample 5 different correct solutions. Then, we filter the samples with text length over 1024. Finally, we get 56k instances for fine-tuning.

To make it easier for the model to distinguish between task descriptions and solutions, we format training instances for fine-tuning. For these instances in MBPP, we concatenate function description and three test cases to form prompt, and then add a <comment> token to the head of the prompt and a <python> token to the end of the prompt. Function code is appended to the <python> token and the <EOT> token is add to the end of function code. Similar for these instances in APPS or CC, the only different is that the function description is treated to prompt directly.

\begin{figure*}[!ht]
    \centering
    \includegraphics[width=0.8\textwidth]{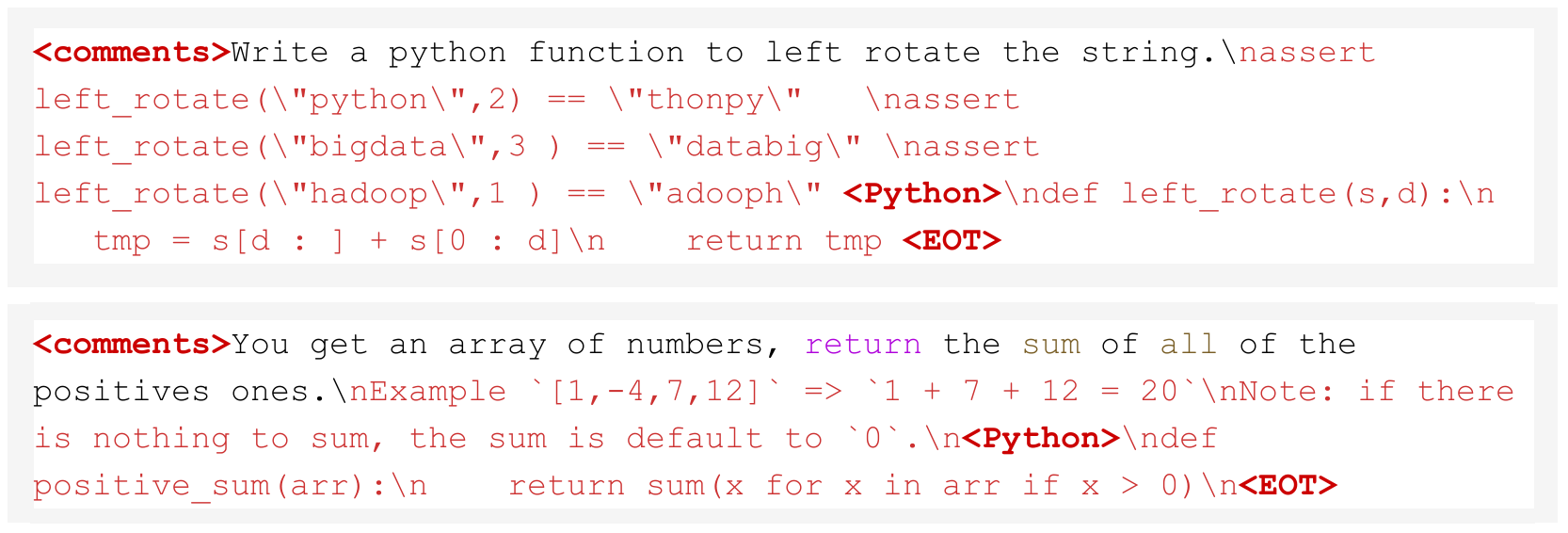}
    \caption{The traning sample format for fine-tuning in MBPP task}
    \label{fig:finetune_sample_2}
\end{figure*}

All these formatted instances from the MBPP fine-tune dataset, APPS and CC constitute the fine-tune datasets. We fine-tune the code domain model, which is extracted from \MODEL\ , for 5 epochs on the fine-tuning datasets.

\subsubsection{Results}

For all sample in MBPP test dataset, function descriptions are augmented with three test cases as prompt, which is similar to the data format used during fine-tuning. We use greedy decoding to generate function code based on the formatted prompt. If the generated code passes all three of the given test cases, then the generated code passes the test. To evaluate the performance of the fine-tuned code domain mode of \MODEL\ , we use the pass@1 as the estimator. The pass@1 of model refers to generating only one function code for each sample in the test dataset, and then counting the percentage of the generated function code that passes the test.

\begin{figure*}[!ht]
    \centering
    \includegraphics[width=0.8\textwidth]{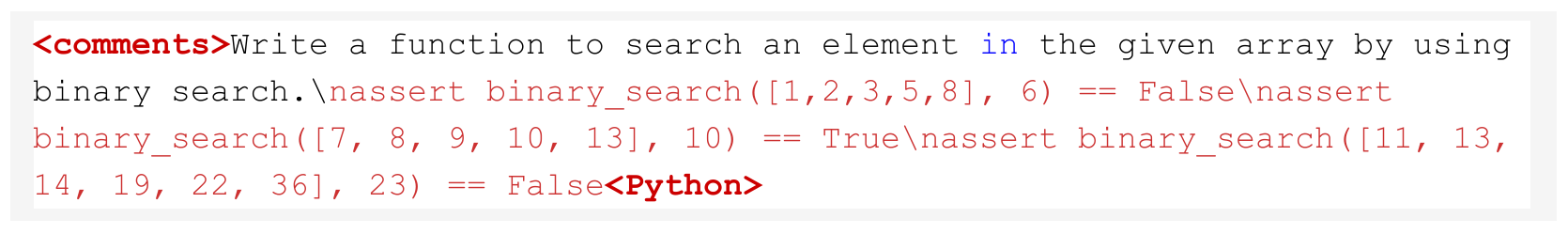}
    \caption{The formatted prompt for generating function code}
    \label{fig:test_prompt}
\end{figure*}

Table~\ref{tab:code_results} shows the comparison of existing models, as well as \MODEL\ on the MBPP dataset, along with model size and number of tokens trained by model. The \MODEL\ outperforms the current state-of-the-art model PanGu-Coder by 1.4 point on the pass@1 for MBPP tasks. The training data of \MODEL\ is less than PanGu-coder, which contains only 75B code data, while Python code data related to MBPP tasks is only 50B data. This suggests that \MODEL\ makes more efficient use of data.

\begin{table}
\centering
\caption{Pass@1 rates on the MBPP dataset, among various models}
\label{tab:code_results}
\begin{tabular}{cclc} 
\hline
Models               & \begin{tabular}[c]{@{}c@{}}Size\\(Billion)~ ~~\end{tabular} & \multicolumn{1}{c}{\begin{tabular}[c]{@{}c@{}}Train tokens \\(Billion)~ ~~\end{tabular}} & \begin{tabular}[c]{@{}c@{}}MBPP(\%)\\PASS@1~ ~~\end{tabular}  \\ 
\hline
INCODER~\cite{Fried2022InCoderAG}              & 6.7B                                                        & 216B (52B python code, 107B other code, other 57)                                        & 19.4                                                          \\
LaMDA~\cite{Thoppilan2022LaMDALM}                & 137B                                                        & 2877B~ ~ ~                                                                               & 14.8                                                          \\
PanGu-Coder~\cite{Christopoulou2022PanGuCoderPS}          & 2.6B                                                        & 387B (all is python code)                                                                & 25.4                                                          \\
\textbf{\MODEL} & 38B                                                         & 300B(50B python code, 25B other code, other 225B)                                        & \textbf{26.8}                                                 \\
\hline
\end{tabular}
\end{table}

\subsection{English Natural Language Understanding}

In order to compare with other large language models on English tasks, we evaluate \MODEL\ model on the SuperGLUE benchmark \cite{wang2019superglue}. SuperGLUE consists of 8 natural language understanding tasks. We use accuracy as the performance metric except for MultiRC dataset where ${\rm F1}$-score over the set of answer options is used (denoted by ${\rm F1}_{a}$). We cast each task to a multiple-choice classification problem. The prediction is chosen based on the maximum log-likelihood score, $\log {\rm P}(completion\,\vert\, context)$, of each available completion given the context. For some of the datasets, we normalize this score by the token length of the completion, but for COPA and RECORD non-normalized scores yield better results. We generally view binary classification in such a way that the completion options are ``Yes'' and ``No'', except for the COPA for which the model chooses between two appropriate sentence continuations. In the table~\ref{en_zeroshot_13B}, we report model's performance on each of the SuperGLUE datasets along with the average score. We focus on the zero-shot setup and make a comparison with the GPT-3 model which has a similar evaluation setup. 
\begin{table}[ht]
\caption{\label{en_zeroshot_13B}
Zero-shot results of English downstream tasks
}
\centering
\begin{tabular}{|c|c|c|c|}
\hline
Dataset   & Metric  & GPT3 13B          & \textbf{PanGu-$\Sigma$} \\ \hline
BoolQ     & acc     & \textbf{66.2}     & 65.54                   \\
CB        & acc     & 19.6              & \textbf{55.36}          \\
Copa      & acc     & \textbf{84.0}     & 79.00                   \\
RTE       & acc     & \textbf{62.8}     & 59.21                   \\
WiC       & acc     & 0.0               & \textbf{50.78}          \\
WSC       & acc     & \textbf{64.4}     & 63.46                   \\
MultiRC   & {\rm $F1_a$}     & \textbf{71.4}     & 59.31                   \\
ReCoRD    & acc     & \textbf{89.0}     & 84.37                   \\ \hline
SuperGLUE & average & 57.2              & \textbf{64.62}          \\ \hline
\end{tabular}
\end{table}

In the Table~\ref{en_zeroshot_13B}, evaluation results are presented. We see that, even with only 112B English tokens, the performance of \MODEL\ English sub-model with 38B parameters roughly meets the performance of the GPT-3 13B model and gets a higher average score.

%% file: sections/Future.tex
In this work, we have present a trillion parameters language model architecture \MODEL\ . With the Random Routed Experts (RRE) and Expert Computation Storage Separation (ECSS), \MODEL\ achieves high system performance under the MindSpore framework using Ascend 910 AI accelerators. By extending and continually training from PanGu-$\alpha$ with 329B tokens, \MODEL\ has successfully achieved state-of-the-art results in a bunch of downstream tasks such as few-shot NLU, open-domain dialogue, question answering, machine translation, and code generation. Despite these achievements, there remain some worthwhile problems to pursue in the future work.
\begin{itemize}
    \item Sparse models offer the benefits of a larger model size at reduced computation cost. Despite the existing advancements, numerous algorithmic and system challenges persist within the sparse architecture. Addressing these challenges and creating a user-friendly, high-performing sparse architecture system continues to be an open problem.
    \item Large language models are designed to be applied in real-world scenarios. Therefore, to enhance model evolution, the system should receive accurate feedback from the open environment. Although InstructGPT~\cite{2022TrainingChat} and ChatGPT~\footnote{\url{https://openai.com/blog/chatgpt}} provide promising approaches, they require a substantial amount of data labeling, which can be time-consuming and costly. Consequently, devising an efficient method to generate valuable signals that can align with the real world is a crucial research topic worth exploring.
   \item Large-scale language models provide intelligent foundations and various modalities alignment objectives for artificial intelligence systems. Therefore, utilizing language models as a foundation and incorporating multiple modalities for perception input in a multimodal model will be one of the most important topics, as already demonstrated by Flamingo~\cite{Alayrac2022FlamingoAV} and GPT-4~\cite{gpt-4} .
    \item Large language models have great potential for real time applications, but their deployment cost remains a major hurdle to overcome. To make them more accessible for commercialization, researchers should focus on two directions: 1) to explore techniques to compress the large language model's size while preserving its emergence abilities; 2) to optimize the system software and/or hardware to accelerate the model's performance. Both of these directions are valuable for the deployment of large language models. 
    \item Online knowledge updates are also critical for optimal performance of the large language model system. However, effectively storing and updating knowledge online is a significant challenge that requires advanced system infrastructure and algorithms. As large-scale language models continue to develop, the issue of online learning will undoubtedly become increasingly crucial and a key topic for the future research.
    
\end{itemize}

%% file: appendices/multiple_choice.tex
\newpage
\clearpage

\setcounter{figure}{0}
\setcounter{table}{0}

\section{Natural Language Generation Examples}
\subsection{Self-chat of dialog\label{Self-chat-appendix}}

\begin{figure*}[!ht]
 \centering
  \centering
    \includegraphics[width=0.77\textwidth]{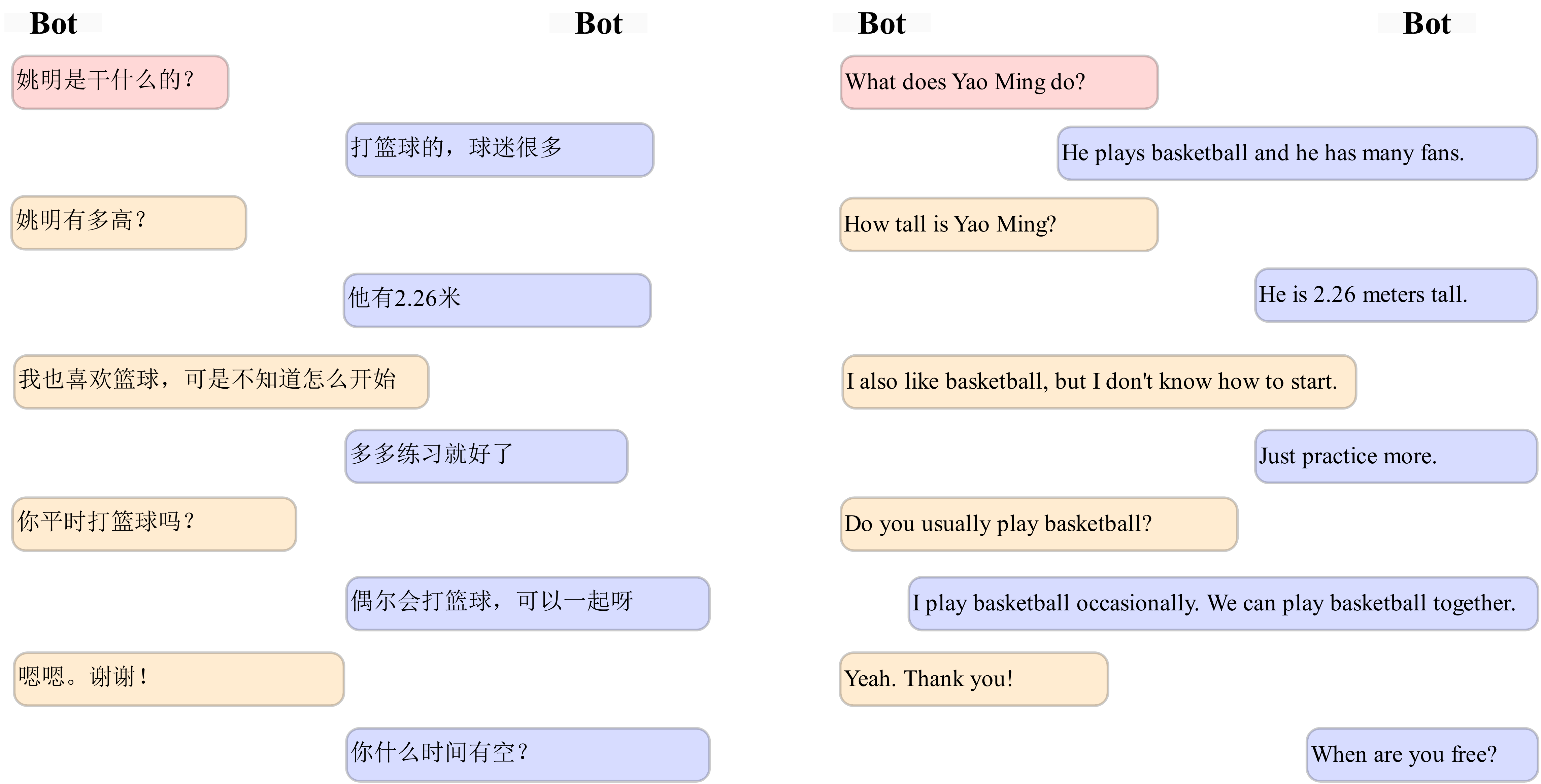}\\
     \includegraphics[width=0.77\textwidth]{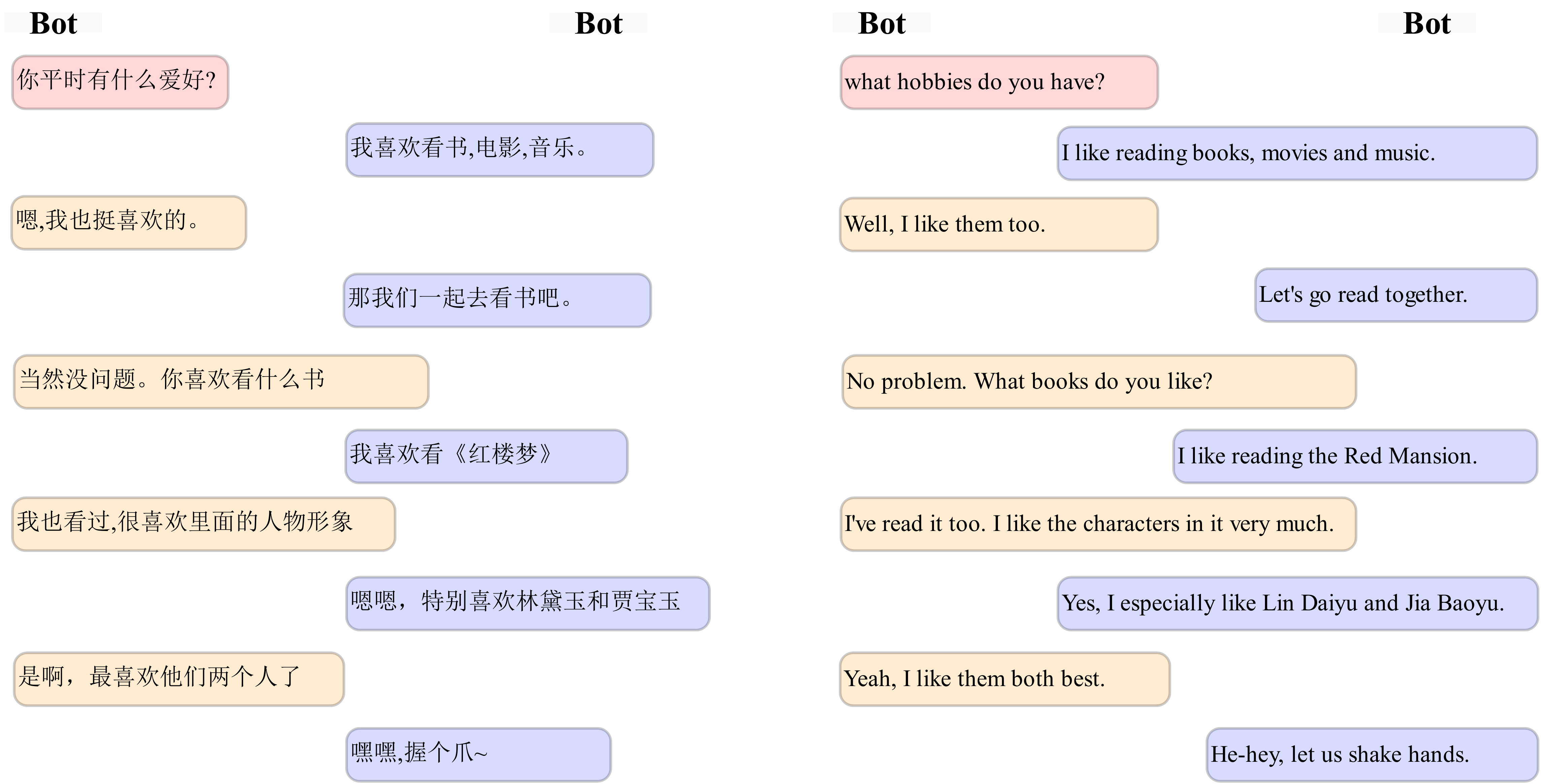}\\
      \includegraphics[width=0.77\textwidth]{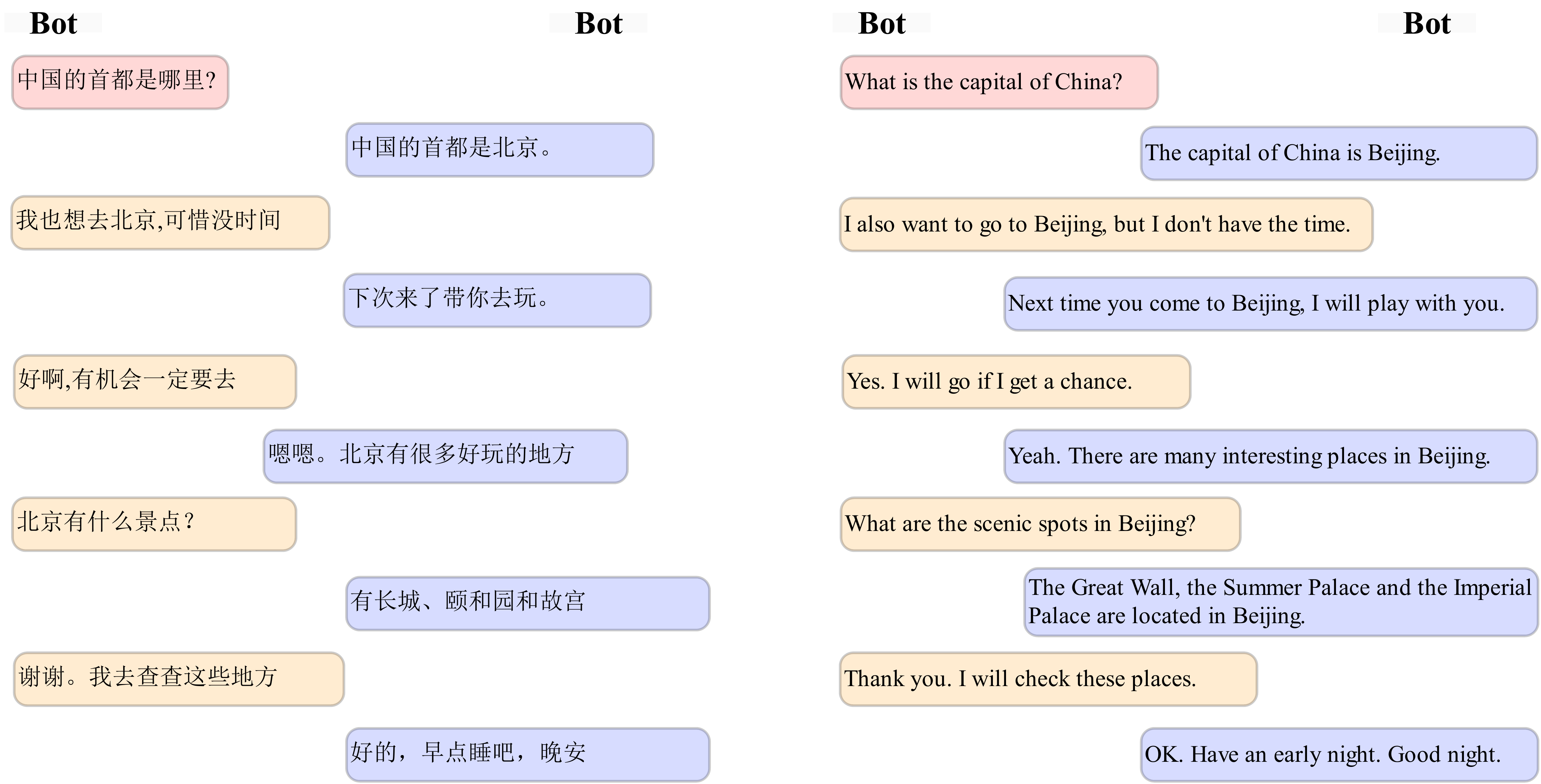}
    \caption{Cases of self-chat
    }
    \label{fig:selfchat2}
\end{figure*}

\subsection{Topic-grounded dialog generation
\label{chinese_dialogue_case_topicground-appendix}}

\begin{CJK*}{UTF8}{gbsn}
\begin{table}[h]
\caption{Cases of topic-grounded dialog}
\begin{center}
\begin{tabular}{ll}
\hline
Dialog history & \parbox[c]{13cm}{你知道,今年的世预赛吗？马尔代夫那个     \\
(Do you know the World Cup Qualification held in Maldives?)\\
知道知道,没想到国足客场居然5-0大胜马尔代夫,难以置信。\\
(Yes, it's unbelievable that the China Men's National Football Team beat Maldives 5-0 away.)\\
对啊，打马尔代夫的池忠国都进化成坎特了!     \\
(Yeah, Chi Zhongguo evolved into Kant when playing with Maldives team.)\\
有那么夸张吗？或者对手实在太菜了才赢的？  \\
(Isn't it so dramatic? Or it is because the opponent is too weak?)\\
对手确实有点菜，所以才能让里皮在人员和战术上有很多尝试。\\
(The opponent is very weak exactly. Therefore, Lippi can make many attempts on personnel rotation and tactics.)
}\\
 \cline{2-2}
 EVA2.0  &  \parbox[c]{13cm}{{\color[HTML]{333333} \textbf{所以我觉得应该让马塞洛打后卫。}\\
 (That's why I think Marcelo should play fullback.) } } \\
\cline{2-2}
PanGu-Bot 2.6B  & \parbox[c]{13cm}{ {\color[HTML]{333333} \textbf{也是,毕竟国足教练也是很强}}\\
(Yes, after all, the coach of the China Men's National Football Team is also very strong.)
}\\
\cline{2-2}
\textbf{PanGu-$\Sigma$}  & \parbox[c]{13cm}{ {\color[HTML]{333333} \textbf{对啊，银狐里皮的战术变化多端。}}\\
(Yes, the tactics of "Silver Fox" Lippi are changeable.)
}\\
\hline
\hline
Dialog history & \parbox[c]{13cm}{{\color[HTML]{333333} \textbf{你也在看这篇报道啊?}}     \\
(Are you also reading this report?)\\
          {\color[HTML]{333333} \textbf{嗯,刚刷到就点进来看看。}}  \\
          (Yes, I click on the link after I saw the web page.)\\
 {\color[HTML]{333333} \textbf{你喜欢拳王泰森嘛?}}     \\
 (Do you like the boxing champion Tyson?)\\
  {\color[HTML]{333333} \textbf{还好吧,觉得他还是挺厉害的,虽然我更喜欢拳王阿里。真男人,真心帅。}}  \\
  (Not bad. I think he's pretty good. I like boxing champion Ali more, a real man and handsome.)\\
  {\color[HTML]{333333} \textbf{真的吗?我对这方面不太了解,你可以多跟我说说拳王阿里吗?}}  \\
  (Really? I don't know much about this. Can you tell me more about boxing champion Ali?)
  } \\
   \cline{2-2}
 EVA2.0  &  \parbox[c]{13cm}{{\color[HTML]{333333} \textbf{我也是看了泰森的报道才知道的。}\\
 (I found out after reading Tyson's story.) } } \\
  \cline{2-2}
PanGu-Bot 2.6B  & \parbox[c]{13cm}{ {\color[HTML]{333333} \textbf{好的,我会多和你说说的。}}\\
(Okay, I'll talk to you more.)
}\\
  \cline{2-2}
\textbf{PanGu-$\Sigma$}  &  \parbox[c]{13cm}{ {\color[HTML]{333333} \textbf{阿里的影响力远远超出了体育的范畴。}} \\
(Ali's influence goes far beyond the scope of sporting competitions.)
}\\

\hline
\hline
Dialog history & \parbox[c]{13cm}{
{\color[HTML]{333333} \textbf{今天天气不错是吧!}}     \\
(It's a nice day today.)\\
          {\color[HTML]{333333} \textbf{没错,非常适合出来散散心。}}  \\
          (Yes, it is the time to go out for relaxation.)\\
 {\color[HTML]{333333} \textbf{CBA的新赛季马上开始了,北控今天进行了誓师大会。}}     \\
 (The new season of CBA will begin soon, and Beikong held the swearing ceremony today.)\\
  {\color[HTML]{333333} \textbf{誓师大会会鼓舞队员的士气。}}  \\
  (The swearing ceremony can inspire the morale of the team.)\\
  {\color[HTML]{333333} \textbf{没错!新赛季状元秀王少杰也要跟随北控进入新赛季的比赛。}}  \\
  (Yes, Wang Shaojie, the first-pick of the draft in the new season, will also participate in competition as a member of Beikong.)
  } \\
   \cline{2-2}
 EVA2.0  &  \parbox[c]{13cm}{{\color[HTML]{333333} \textbf{确实是！}\\
 (Indeed it is!) } } \\
  \cline{2-2}
PanGu-Bot 2.6B  & \parbox[c]{13cm}{ {\color[HTML]{333333} \textbf{这就是誓师大会的魅力所在。}}\\
(This is the charm of the swearing ceremony.)
} \\
\cline{2-2}
\textbf{PanGu-$\Sigma$}  & \parbox[c]{13cm}{ {\color[HTML]{333333} \textbf{王少杰的实力不容小视啊,他的投篮姿势很帅气。}}\\
(Wang Shaojie's strength cannot be undervalued, and his shooting form is very beautiful.)
}
\\
\hline

\end{tabular}
\end{center}
\label{chinese_dialogue_case}
\end{table}
\end{CJK*}



\clearpage

\subsection{Open domain question-answering of dialog model\label{chinese_dialogue_case_qa-appendix}}

\begin{CJK*}{UTF8}{gbsn}
\begin{table}[h]
\caption{Cases of open domain question-answering}
\begin{center}
\begin{tabular}{lll}
\hline
Question: & \parbox[c]{6cm}{世界上的四大文明古国是哪四个？}  & \parbox[c]{6cm}{What are the four ancient civilizations in the world?}\\
\cline{2-3}
EVA2.0:&  \parbox[c]{6cm}{世界上的四大文明古国:世界上的四大文明古国} & \parbox[c]{6cm}{The four ancient civilizations in the world: the four ancient civilizations in the world.}
 \\
\cline{2-3}
PanGu-Bot 2.6B:&  \parbox[c]{6cm}{古巴比伦、古印度、古埃及、古埃及} & \parbox[c]{6cm}{Ancient Babylon, Ancient India, Ancient Egypt and Ancient Egypt}\\
\cline{2-3}
\textbf{PanGu-$\Sigma$}:&  \parbox[c]{6cm}{古印度、古埃及、古巴比伦、中国} & \parbox[c]{6cm}{Ancient India, Ancient Egypt, Ancient Babylon and China}
\\
\hline
Question: & \parbox[c]{6cm}{目前，中国的人口大约有多少亿？}     & \parbox[c]{6cm}{At present, how many billions of people are there in China?}\\
\cline{2-3}
EVA2.0:&  \parbox[c]{6cm}{我觉得应该是100亿左右} & \parbox[c]{6cm}{I think it's about 10 billion.}
 \\
\cline{2-3}
PanGu-Bot 2.6B:&  \parbox[c]{6cm}{960万,占全国总人口的百分之七十。} & \parbox[c]{6cm}{9.6 million, 70 percent of China's total population.} \\
\cline{2-3}
\textbf{PanGu-$\Sigma$}:&  \parbox[c]{6cm}{中国有十四亿人口} & \parbox[c]{6cm}{China has a population of 1.4 billion.}
\\
\hline
Question: & \parbox[c]{6cm}{美术中的“三原色”指哪三种颜色}     & \parbox[c]{6cm}{What are the "three primary colors" in art?}\\
\cline{2-3}
EVA2.0:& \parbox[c]{6cm}{三原色指的是三原色的颜色,指的是三原色的颜色。} & \parbox[c]{6cm}{Yellow, red and green}
 \\
\cline{2-3}
PanGu-Bot 2.6B:&  \parbox[c]{6cm}{红、黄、蓝、绿、紫} & \parbox[c]{6cm}{Red, yellow, blue, green, purple.} \\
\cline{2-3}
\textbf{PanGu-$\Sigma$}:&  \parbox[c]{6cm}{红、黄、蓝} & \parbox[c]{6cm}{Red, yellow and blue.}
\\
\hline
\end{tabular}
\end{center}
\label{chinese_dialogue_case}
\end{table}
\end{CJK*}

\clearpage

\subsection{Character dialog generation\label{chinese_dialogue_case_character-appendix}}
\begin{CJK*}{UTF8}{gbsn}
\begin{table}[h]
\caption{A case of character dialog. Source of the prompt about 姚明(Yaoming) is from https://baike.baidu.com. }
\begin{center}
\begin{tabular}{ll}
\hline 
Prompt &  \parbox[c]{13cm}{ 姚明,前中国职业篮球运动员,司职中锋,现任亚洲篮球联合会主席、中国篮球协会主席。在2002年NBA选秀中,他以状元秀身份被NBA的休斯敦火箭队选中,2003-09年连续6个赛季(生涯共8次)入选NBA全明星赛阵容。\\
(Yao Ming, a former Chinese professional basketball player, working as a center, is currently president of the Asian Basketball Federation and president of the Chinese Basketball Association. In the 2002 NBA Draft, he was selected by the NBA's Houston Rockets as the top draft, and was selected for the NBA All-Star Game for six consecutive seasons (eight career times) from 2003 to 2009.)\\
姚明:你好,我是姚明,NBA球员,现在是中国篮协主席。\\
(Yao Ming: Hello, I'm Yao Ming, NBA player, now the president of the Chinese Basketball Association.)\\
记者:你好,非常高兴今天和你对话。你认为中国篮球应该如何改革?\\
(Reporter: Hello, I'm very happy to talk to you today. How do you think Chinese basketball should be reformed?)
}  \\
\cline{2-2}
\textbf{PanGu-$\Sigma$} &   \parbox[c]{13cm}{
姚明:首先,我觉得中国篮球的改革,首先要从体制上来改革,中国篮球在过去的几十年,特别是改革开放这几十年,取得了很大的成就,但在中国篮球的体制上,还是有很多的问题,所以中国篮球的改革,首先要从体制上来改革。\\
(Yao Ming: First of all, I think the system of China's basketball must first be reformed. China's basketball has made great achievements in the past decades, especially in the decades of reform and opening up. However, there are still many problems in the system of China's basketball. Therefore, Chinese basketball must first be reformed from the system.)\\
记者:我想问一个问题,你在NBA的时候,你的队友都叫你“小巨人”,你觉得在中国的球迷当中,你的绰号是什么?\\
(Reporter: I want to ask a question. When you were in the NBA, your teammates called you "Little Giant". What do you think is your nickname in the Chinese fans?)
\\
姚明:其实在中国,大家都叫我“小巨人”,我觉得这是对我的一种认可,我希望自己能够成为中国篮球的“小巨人”。\\
(Yao Ming: In fact, in China, everyone calls me "little giant", and I think this is an approval for me. I hope that I can become the "little giant" of Chinese basketball.)\\
记者:我知道你是中国篮球的标志性人物,你对中国篮球的未来有什么展望?\\
(Reporter: I know you are the iconic figure of Chinese basketball. What are your prospects for the future of Chinese basketball?)\\
姚明:我觉得中国篮球在未来的发展当中,应该更多的去重视青少年的培养,因为中国篮球的未来,还是要靠青少年来培养,只有青少年的发展了,中国篮球才会有更多的未来。\\
(Yao Ming: I think in the future development of Chinese basketball, we should pay more attention to the training of young people. Because Chinese basketball will have more future with relying on the development of young people.)\\
记者:你的球迷也都非常关心你的伤病,在伤病的恢复过程当中,你都做了哪些康复训练?\\
(Reporter: Your fans are also very concerned about your injury. In the recovery process, what have you done on rehabilitation training?)\\
姚明:其实在康复训练方面,我是比较注重的,因为我的伤病,也是比较多的。\\
(Yao Ming: In fact, I paid more attention to the rehabilitation training, because I had a lot of injuries.)
}\\
\hline
\end{tabular}
\end{center}
\label{chinese_dialogue_case}
\end{table}
\end{CJK*}

\begin{CJK*}{UTF8}{gbsn}
\begin{table*}[]
\caption{A case of character dialog. Source of the prompt about 马斯克(Musk) is from https://baike.baidu.com.}
\begin{center}
\begin{tabular}{ll}
\hline 
 Prompt &  \parbox[c]{13cm}{ 埃隆·里夫·马斯克,1971年6月28日出生于南非的行政首都比勒陀利亚,企业家、工程师、慈善家、美国国家工程院院士。2021年3月2日,胡润研究院发布《2021胡润全球富豪榜》,马斯克以1.28万亿元人民币财富首次成为世界首富,他是特斯拉公司创始人。\\
 (Elon Reeve Musk was born on June 28, 1971 in Pretoria, the administrative capital of South Africa. He is an entrepreneur, engineer, philanthropist and member of the National Academy of Engineering. On March 2, 2021, Hurun Research Institute released the "2021 Hurun Global Rich List", and Musk became the world's richest man for the first time with a wealth of 1.28 trillion yuan. In addition, he is the founder of Tesla.)
\\
马斯克:你好,我是马斯克,特斯拉公司创始人。
\\
(Musk: Hello, I'm Musk, founder of Tesla.)\\
记者:你好,非常高兴今天和你对话。你怎么看待可持续能源的发展?\\
(Reporter: Hello, it's a pleasure to talk to you today. What do you think of the development of sustainable energy?)
}  \\
\cline{2-2}
\textbf{PanGu-$\Sigma$} &   \parbox[c]{13cm}{
马斯克:可持续能源是一个非常重要的话题,我认为我们需要做的是通过技术的进步,来改变我们的能源结构。可持续能源的发展是一个非常重要的趋势,因为我们现在的能源结构是非常不合理的。我们需要找到一种方式,让能源的使用更加清洁,更加高效,并且更加环保。\\
(Musk: Sustainable energy is a very important topic, and I think what we need to do is to change our energy mix through technological advances. The development of sustainable energy is a very important trend, because our current energy structure is very unreasonable. We need to find a way to make the use of energy cleaner, more efficient and more environmentally friendly.)

记者:你觉得特斯拉的成功有哪些秘诀?\\
(Reporter: What do you think is the secret to Tesla's success?)
\\
马斯克:我觉得最重要的一点是,我们有一个非常好的团队。特斯拉的员工是世界上最聪明的一群人,他们为特斯拉的成功付出了很多努力。我认为,一个公司的成功取决于它的团队,而不是它的产品。\\
(Musk: I think the most important thing is that we have a very good team. Tesla's employees are some of the smartest people in the world, and they've worked so hard for Tesla's success. I think the success of a company depends on its team, not on its products.)\\
记者:你认为特斯拉的产品是如何影响世界的?\\
(Reporter: How do you think Tesla's products have impacted the world?)
\\
马斯克:特斯拉的产品正在改变世界,因为我们的产品可以让人们的生活变得更美好。我们的产品可以让人们的出行变得更便宜,让人们的生活变得更环保。\\
(Musk: Tesla products are changing the world, because our products can make people's lives better. Our products can make people travel cheaper and make people's lives more environmentally friendly.)\\
记者:你觉得特斯拉的成功对世界意味着什么?\\
(Reporter: What do you think Tesla's success means to the world?)\\
马斯克:特斯拉的成功让很多人看到了新的希望,让很多人看到了未来的发展方向。我们的成功也让更多的人看到了创业的机会,让更多的人有机会去改变世界。
\\
(Musk: The success of Tesla let a lot of people see new hope, and let a lot of people see the future development direction. Our success has also allowed more people to see the opportunity to start a business and change the world.)
}
   \\
\hline
\end{tabular}

\end{center}
\label{chinese_dialogue_case}
\end{table*}
\end{CJK*}

\begin{CJK*}{UTF8}{gbsn}
\begin{table*}[]
\caption{A case of character dialog. Source of the prompt about 乔布斯(Jobs) is from https://baike.baidu.com.}
\begin{center}
\begin{tabular}{ll}
\hline
Prompt &  \parbox[c]{13cm}{ 史蒂夫·乔布斯是一名美国发明家、企业家、营销家,苹果公司联合创始人之一,曾任董事长、首席执行官职位,NeXT创办人及首席执行官,也是皮克斯动画创办人并曾任首席执行官,2006年为华特迪士尼公司董事会成员。乔布斯在1970年代末个人电脑的热潮中创业,与苹果公司另一始创人斯蒂夫·沃兹尼亚克及首任投资者迈克·马库拉协同其他人设计、开发及销售AppleII系列。\\
(Steve Jobs is an American inventor, entrepreneur, marketer, co-founder of Apple, former chairman and CEO, founder and CEO of NeXT, founder and former CEO of Pixar Animation, and was a member of the Walt Disney Company's board of directors in 2006. Jobs started his business in the late 1970s during the personal computer boom, working with Apple's other founder Steve Wozniak and first investor Mike Makula to design, develop and sell the Apple II series.)
\\
 乔布斯:你好,我是乔布斯,苹果公司联合创始人之一。
 \\
(Jobs: Hello, I'm Jobs, co-founder of Apple.)\\
记者:你好,非常高兴今天和你对话。你对未来手机行业的发展怎么看?\\
(Reporter:Hello, it's a pleasure to talk to you today. What do you think of the future development of the mobile phone industry?)
}  \\
\cline{2-2}
\textbf{PanGu-$\Sigma$} &   \parbox[c]{13cm}{
乔布斯:手机行业未来肯定会有更多的创新,因为手机的功能越来越多,可以做的事情也越来越多。但是,我认为手机行业的创新会越来越集中在高端市场,比如,像苹果这样的公司,会推出一些高端的产品,而其他的手机厂商,可能会在中低端市场上进行创新。\\
(Jobs: There will certainly be more innovation in the mobile phone industry in the future, because mobile phones have more and more functions and are capable to do more things. However, I think innovation in the mobile phone industry will become more and more concentrated in the high-end market. For example, companies like Apple will launch some high-end products, while other mobile phone manufacturers may innovate in the mid-range and low-end market.)
\\
记者:在智能手机领域,中国的发展速度是很快的,你觉得中国的手机行业会有哪些新的趋势?\\
(Reporter: In the smartphone field, China is developing rapidly. What new trends do you think will happen in China's mobile phone industry?)
\\
乔布斯:我觉得中国的手机行业发展得非常好,中国的手机行业在未来的发展中,也会不断地创新,不断地推出新的产品。\\
(Jobs: I think China's mobile phone industry is developing very well. In the future, China's mobile phone industry will continue to innovate and launch new products.)
\\
记者:在智能手机的市场上,苹果公司和三星公司一直都是竞争对手,你觉得在竞争激烈的市场上,三星公司和苹果公司谁会更胜一筹?\\
(Reporter: Apple and Samsung have always been rivals in the smartphone market. For Samsung or Apple, who do you think is better in a competitive market?)
\\
乔布斯:我觉得在竞争激烈的市场上,竞争对手之间的竞争,主要是看谁能够推出更多的新产品,谁能够在产品的创新上更有优势。\\
(Jobs: I think in a competitive market, the competition between competitors is mainly about who can launch more new products and who can have a better advantage in product innovation.)\\
记者:苹果公司在中国的发展速度非常快,中国也成为了苹果公司最重要的市场之一,在你看来,苹果公司在中国市场上的发展,对全球的市场发展有什么影响?\\
(Reporter:Apple's development in China is very fast, and China has become one of Apple's most important markets. In your opinion, what impact does Apple's development in the Chinese market have on the development of the global market?)
\\
乔布斯:我觉得苹果公司在中国市场上的发展,对全球的市场发展,肯定会有非常大的影响。因为中国的市场非常大,中国的消费者也非常多。
\\
(Jobs: I think Apple's development in the Chinese market will certainly have a great impact on the development of the global market. Because China's market is very big, and China has a lot of consumers.)
}\\
\hline
\end{tabular}

\end{center}
\label{chinese_dialogue_case0}
\end{table*}
\end{CJK*}

\clearpage

\subsection{Open-ended text generation\label{chinese_dialogue_case_longQA-appendix}}

\begin{CJK*}{UTF8}{gbsn}
\begin{table}[h]
\caption{Cases of reading comprehension. Source of the passages are from https://baike.baidu.com.}
\begin{center}
\begin{tabular}{ll}
\hline
Passage & \parbox[c]{13cm}{阅读文章:鲁迅,原名周樟寿,后改名周树人,浙江绍兴人。著名文学家、思想家、革命家、教育家、民主战士,新文化运动的重要参与者,中国现代文学的奠基人之一。
早年与厉绥之和钱均夫同赴日本公费留学,于日本仙台医科专门学校肄业。鲁迅,1918年发表《狂人日记》时所用的笔名,也是最为广泛的笔名。鲁迅一生在文学创作、文学批评、思想研究、文学史研究、翻译、美术理论引进、基础科学介绍和古籍校勘与研究等多个领域具有重大贡献。他对于五四运动以后的中国社会思想文化发展具有重大影响,蜚声世界文坛,尤其在韩国、日本思想文化领域有极其重要的地位和影响,被誉为二十世纪东亚文化地图上占最大领土的作家”。请问:鲁迅于1918年发表过什么文章?回答:\\
(Read the article: Lu Xun, formerly known as Zhou Zhangshou, later renamed Zhou Shuren, Zhejiang Shaoxing. He is a famous writer, thinker, revolutionist, educator, democracy fighter, important participant of the New Culture Movement, one of the founders of Chinese modern literature. 
In his early years, Lu Xun, Li Suizhi and Qian Junfu went to Japan to study abroad at public expense and studied at Sendai Medical School in Japan. Lu Xun, the pen name used when he published Diary of Mad Men in 1918, is also the most widely used pen name. Lu Xun made great contributions in literary creation, literary criticism, ideological research, literary history research, translation, introduction of art theory, introduction of basic science and collation and research of ancient books. He had a great influence on the ideological and cultural development of Chinese society after the May 4th Movement, and was famous in the world literary circles, especially in the ideological and cultural fields of Korea and Japan. He was known as the writer who occupied the largest territory on the cultural map of East Asia in the 20th century. Question: What articles did Lu Xun publish in 1918? Answer:)
}  \\
\cline{2-2}
\textbf{PanGu-$\Sigma$} & \parbox[c]{13cm}{鲁迅于1918年发表了《狂人日记》。《狂人日记》是中国现代文学史上最早的一部白话小说,也是中国现代小说的开山之作。\\(Lu Xun published the Madman's Diary in 1918. Madman's Diary is the earliest vernacular novel in the history of modern Chinese literature, and is also the beginning of modern Chinese novels.)
}  \\

\hline
Passage & \parbox[c]{13cm}{阅读文章:深圳,简称深,别称鹏城,广东省辖地级市,是广东省副省级市,国家计划单列市,超大城市,国务院批复确定的中国经济特区、全国性经济中心城市、国际化城市、科技创新中心、区域金融中心、商贸物流中心。全市下辖9个行政区和1个新区,总面积1997.47平方千米。截至2021年末,深圳市常住人口1768.16万人。1980年,成为中国设立的第一个经济特区,中国改革开放的窗口和新兴移民城市,创造了举世瞩目的深圳速度。请问:深圳的别称是什么?回答:\\
(Read the article: Shenzhen, also known as Pengcheng, is a prefecture-level city under the jurisdiction of Guangdong Province. It is a sub-provincial city in Guangdong Province, a city separately listed on the national plan and a super-large city. China's special economic zones, national economic centers, international cities, scientific and technological innovation centers, regional financial centers and trade logistics centers approved by the State Council. The city has 9 administrative districts and 1 new district under its jurisdiction, with a total area of 1997.47 square kilometers. By the end of 2021, Shenzhen had a resident population of 17.6816 million. In 1980, it became the first special economic zone established by China, the window of China's reform and opening up and the emerging immigrant city, creating a speed of Shenzhen that attracted worldwide attention. Question: What is the nickname of Shenzhen? Answer:)
}  \\
\cline{2-2}
\textbf{PanGu-$\Sigma$} & \parbox[c]{13cm}{深圳别称鹏城\\(Shenzhen is also known as Pengcheng.)
}  \\
\hline
\end{tabular}

\end{center}
\label{chinese_dialogue_case1}
\end{table}
\end{CJK*}

\begin{CJK*}{UTF8}{gbsn}
\begin{table*}[h]
\caption{Cases of QA}
\begin{center}
\begin{tabular}{ll}
\hline

Query & {\makecell[l]{人工智能能干什么?\\(What can artificial intelligence do?)
}}  \\
\cline{2-2}
\textbf{PanGu-$\Sigma$} & \parbox[c]{13cm}{在当今这个时代,AI已经无处不在,从互联网到移动互联网,从电子商务到移动支付,从社交网络到O2O,AI的身影无处不在。而在医疗领域,AI也正逐渐渗透到医疗各个领域,并发挥着越来越重要的作用。在医院的各个科室,我们都能看到AI的身影。在手术室,医生可以通过AI来观察患者的生命体征,及时对患者进行手术;在门诊,医生可以通过AI来对患者的病情进行分析,为患者制定最合适的治疗方案;在病理实验室,医生可以通过AI来对患者的病理进行分析,及时为患者制定最合适的治疗方案。在未来,AI的应用将越来越多,而这些应用的背后,都离不开数据的支持。\\
(In today's era, AI is everywhere, from the Internet to mobile Internet, from e-commerce to mobile payment, from social networks to O2O. In the medical field, AI is gradually infiltrating into the medical field and playing an increasingly important role. In every department of the hospital, we can see AI. In the operating room, doctors can observe the patient's vital signs through AI and operate on the patient in time. In the outpatient department, doctors can use AI to analyze the patient's condition and develop the most appropriate treatment plan for the patient. In the pathology laboratory, doctors can use AI to analyze the patient's pathology and develop the most appropriate treatment plan for the patient in time.
In the future, 
more and more AI applications will be developed, and these applications must be supported by data.)
}  \\

\hline
 Query &  {\makecell[l]{请详细说明熬夜的坏处。\\Please elaborate on the disadvantages of staying up late.}}  \\
 \cline{2-2}
\textbf{PanGu-$\Sigma$} &  \parbox[c]{13cm}{熬夜是现在很多人都有的一种习惯,因为工作的原因,很多人都会选择熬夜,那么,熬夜对身体有什么危害呢?\\
1、导致皮肤受损。
熬夜对皮肤的伤害是非常大的,经常熬夜会让皮肤的水分流失,皮肤的代谢能力也会下降,导致皮肤出现各种问题,比如长痘、黑眼圈、脸色暗沉等。\\
2、导致免疫力下降。
经常熬夜会导致人体的免疫力下降,从而导致各种疾病的发生,尤其是对于女性来说,经常熬夜会导致内分泌失调,从而导致月经不调等问题。\\
3、导致精神萎靡。
经常熬夜会让人精神萎靡,经常会感到疲劳,从而影响到工作和生活,所以经常熬夜的人往往会感到精神不佳,做事无精打采。\\
4、导致肠胃不适。
经常熬夜会导致人的肠胃功能紊乱,出现消化不良的问题,所以经常熬夜的人经常会出现腹胀、腹泻等肠胃不适的症状。\\
5、导致失眠。
经常熬夜会导致人的睡眠质量下降,经常熬夜的人往往会出现失眠的问题,而且失眠的时间会越来越长,严重的甚至会导致人出现神经衰弱的问题。\\
6、导致记忆力下降。
经常熬夜会导致人的记忆力下降,经常熬夜的人往往会出现记忆力减退的问题,而且会出现头晕、头痛等症状。\\
(Staying up late is a habit that many people have now. Because of work, many people will choose to stay up late. So, what is the harm to the health of staying up late?\\
1. Causes skin damage.
Staying up late is very harmful to the skin. Often staying up late will cause the skin's moisture loss, and the skin's metabolic capacity will also decline, which will lead to various skin problems, such as acne, dark circles under the eyes, dullness, etc..\\
2. Causes reduced immunity.
Often staying up late can lead to the decline of human immunity, which can lead to various diseases. Especially for women, often staying up late can lead to endocrine disorders, which can lead to irregular menstruation and other problems.\\
3. Causes mental malaise.
Often staying up late can make people mentally depressed, feel tired, thus affecting work and life. Therefore, often staying up late will cause mental malaise, make people be listless.\\
4. Causes gastrointestinal discomfort.
Often staying up late can lead to gastrointestinal dysfunction and indigestion problems, so people who often stay up late often have abdominal distension, diarrhea and other gastrointestinal discomfort symptoms.\\
5. Causes insomnia.
Often staying up late will lead to the decline of sleep quality, and the time of insomnia will be longer and longer. What's worse, it even can lead to the problem of neurasthenia.\\
6. Causes memory loss.
People who stay up late often will have memory loss problems, and cause symptoms such as dizziness and headache.)
}\\
\hline
\end{tabular}

\end{center}
\label{chinese_dialogue_case2}
\end{table*}
\end{CJK*}

\begin{CJK*}{UTF8}{gbsn}
\begin{table*}[]
\caption{A case of text generation with few-shot prompt learning}
\begin{center}
\begin{tabular}{ll}
\hline 
\textbf{Prompt} & \parbox[c]{13cm}{
上联：马上催征催上马\\
(The first couplet: Expedite the expedition, and urge the horse immediately.)\\
下联：春迎快乐快迎春\\
(The second couplet: Welcome happiness in spring, and welcome spring immediately.)
\\ \\
上联：浮生若梦凭心写\\
(The first couplet: Life is like a dream, which should be written with heart.)\\
下联：此世如歌任意弹\\
(The second couplet: Play freely the life which is like a song.)
\\\\
上联：苍官青士左右树 \\
(The first couplet: Old officials, young men, and trees on the left and right.)
}  \\
\cline{2-2}
\textbf{PanGu-$\Sigma$} & \parbox[c]{13cm}{
下联：海内名花满门香
\\
(The second couplet: The famous flowers in the country are full of fragrance.)
}  \\
\hline
\end{tabular}
\end{center}
\label{chinese_dialogue_case_longQA_fewshot2}
\end{table*}
\end{CJK*}

\begin{CJK*}{UTF8}{gbsn}
\begin{table*}[h]
\caption{A case of coreference resolution with few-shot prompt learning}
\begin{center}
\begin{tabular}{ll}
\hline

\textbf{Prompt} & \parbox[c]{13cm}{
Q：苏小姐听了，心里直刺痛，回答孙太太同时安慰自己道：``那绝不可能！鲍小姐有未婚夫，她自己跟我讲过。她留学的钱还是她未婚夫出的。”\\
``她”指代的内容是孙太太吗？\\
(Q: When Miss Su heard this, her heart ached. She answered Mrs. Sun and comforted herself, "That's impossible! Miss Bao has a fiance, and she told me about it herself. Her money for studying abroad is still from her fiance."\\
Does "she" refer to Mrs. Sun?)\\
A：不是，``她”指代的是鲍小姐\\
(A:No, "she" refers to Miss Bao.)\\

Q：这家人也有一个女孩子，年纪不大，远远地盯着我，害羞而好奇。我一对上她的目光，她就慌不迭扭头而去。\\
上面这句话中的``她”指代的是什么？\\
(Q: There was also a girl in this family. She was young and looked at me from a distance, shy and curious. As soon as I met her eyes, she turned away in a panic.\\
What does "she" refer to in the above sentence?)\\
A：``她”指代的是女孩子\\
(A: "She" refers to a girl.)\\

Q：张先生呵呵大笑，一面分付进来的女佣说：``快去跟太太小姐说，客人来了，请她们出来。”\\
上面这句话中的``她们”指代的是什么？\\
(Q: Mr.Zhang laughed and said to the maid, "Go and tell the ladies and girls that the guests are coming, and ask them out."\\
What does "they" refer to in the above sentence?)
}  \\
\cline{2-2}
\textbf{PanGu-$\Sigma$} & \parbox[c]{13cm}{
A：``她们”指代的是太太小姐,因为太太小姐在屋子里\\
(A: "They" refer to ladies and girls, because they are in the room.)
}  \\
\hline
\end{tabular}

\end{center}
\label{chinese_dialogue_case3}
\end{table*}
\end{CJK*}